\documentclass[10pt,twocolumn,letterpaper]{article}

\usepackage{cvpr}

\usepackage{microtype}
\usepackage[pagebackref,breaklinks,colorlinks,allcolors=cvprblue]{hyperref}
\usepackage[T1]{fontenc}
\usepackage[utf8]{inputenc}
\usepackage{textcomp}
\usepackage{amsmath}
\usepackage{amsfonts}
\usepackage{booktabs}
\usepackage[table]{xcolor}
\usepackage{array}
\usepackage{makecell}
\usepackage{multirow}
\usepackage{adjustbox}
\usepackage{arydshln}
\usepackage{graphicx}
\usepackage{hyperref}
\usepackage{pifont}
\usepackage{fontawesome5}
\usepackage[skins]{tcolorbox}
\usepackage[accsupp]{axessibility}  

\definecolor{cvprblue}{rgb}{0.21,0.49,0.74}
\definecolor{lightblue}{RGB}{230,240,255}

\title{HeartcareGPT: A Unified Multimodal ECG Suite for Dual Signal--Image Modeling and Understanding}

\author{
Yihan Xie\textsuperscript{1}\thanks{Equal contributions.} \quad
Sijing Li\textsuperscript{1}\footnotemark[1] \quad
Zhuonan Wang\textsuperscript{1}\footnotemark[1] \quad
Tianwei Lin\textsuperscript{1}\footnotemark[1] \quad
Chenglin Yang\textsuperscript{1} \quad
Yu Zhong\textsuperscript{1}\\
Wenjie Yan\textsuperscript{1} \quad
Wenqiao Zhang\textsuperscript{1}\thanks{Corresponding author.} \quad
Xiaogang Guo\textsuperscript{1} \quad
Jun Xiao\textsuperscript{1} \quad
Yueting Zhuang\textsuperscript{1} \quad
Beng Chin Ooi\textsuperscript{1} \\
\textsuperscript{1}Zhejiang University \\
{\tt\small \{yihanxie, wenqiaozhang\}@zju.edu.cn}
}

\begin{document}

\maketitle
\begin{abstract}
\label{sec:abstract}

\begin{figure*}[ht]
\centering
\includegraphics[width=0.9\textwidth]{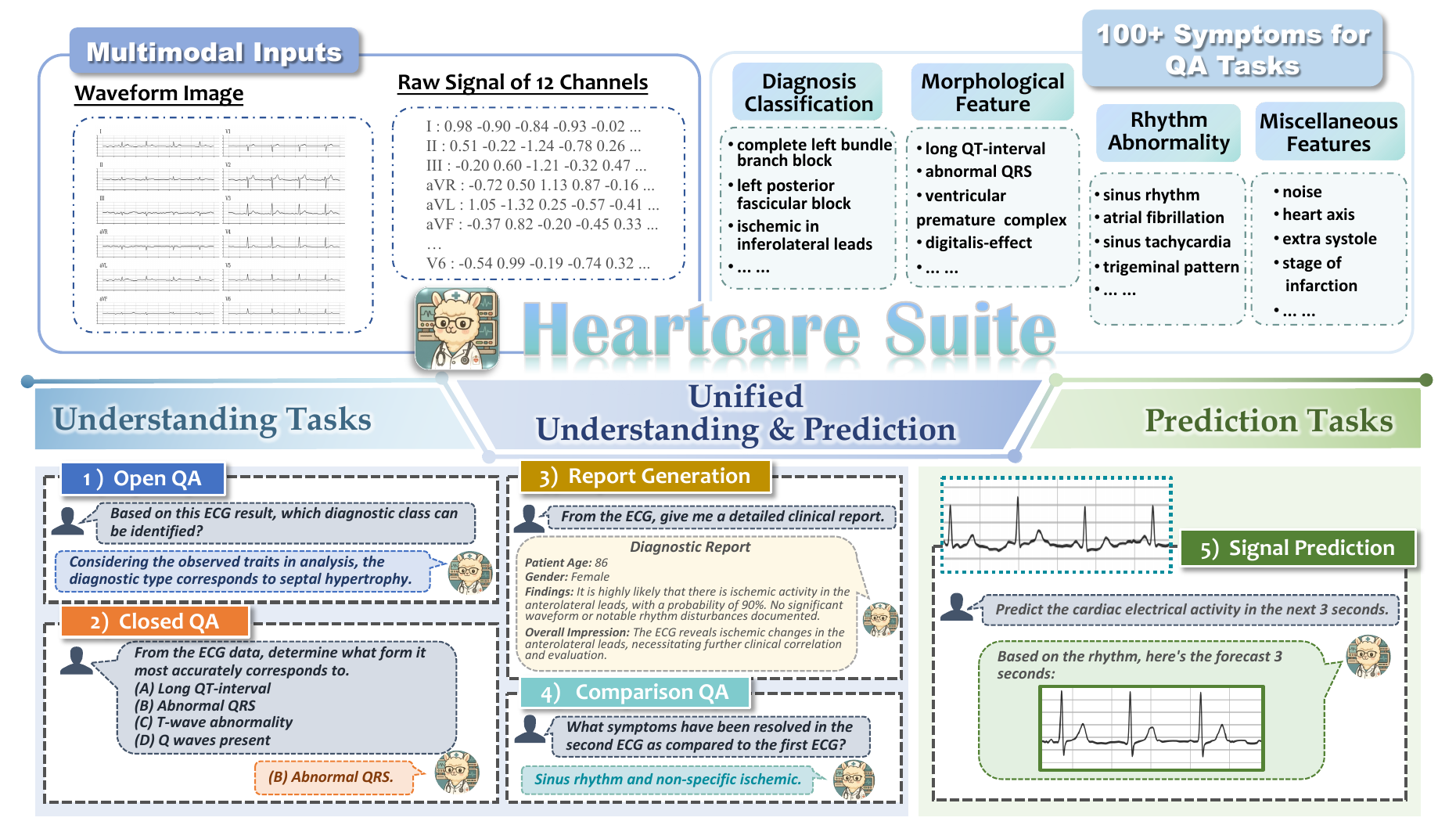}
\vspace{-3mm}
\caption{The proposed \textbf{Heartcare-400K} dataset. Heartcare-400K aggregates real-world ECG data, supporting \textit{Closed-QA}, \textit{Open-QA}, \textit{Comparison-QA}, \textit{Report Generation} and \textit{Signal Prediction}.}
\label{fig:HeartcareGPT}
\vspace{-3mm}
\end{figure*}

Although electrocardiograms (ECG) play a dominant role in cardiovascular diagnosis and treatment, their intrinsic data forms and representational patterns pose significant challenges for medical multimodal large language models (Med-MLLMs) in achieving cross-modal semantic alignment. 
To address this gap, we propose \textbf{Heartcare Suite}, a unified ECG suite designed for dual signal--image modeling and understanding:
\textbf{(i) Heartcare-400K.} A fine-grained ECG instruction dataset on top of our data pipeline engine---\textbf{HeartAgent}---by integrating high quality clinical ECG reports from top hospitals with open-source data.
\textbf{(ii) Heartcare-Bench.} A systematic benchmark assessing performance of models in multi-perspective ECG understanding and cross-modal generalization, providing guidance for optimizing ECG comprehension models.
\textbf{(iii) HeartcareGPT.} Built upon a structure-aware discrete tokenizer \textbf{Beat}, we propose \textbf{Dual Stream Projection Alignment (DSPA)} paradigm---a dual encoder projection alignment mechanism enabling joint optimizing and modeling native ECG signal--image within a shared feature space.
HeartcareGPT achieves consistent improvements across diverse ECG understanding tasks, validating both the effectiveness of the unified modeling paradigm and the necessity of a high-quality data pipeline, and establishing a methodological foundation for extending Med-MLLMs towards physiological signal domains. Our project is available at \url{https://github.com/ZJU4HealthCare/HeartcareGPT}.

\end{abstract}
\section{Introduction}
\label{sec:introduction}

Multimodal large language models (MLLMs)~\cite{gpt4v,liu2023visual,bai2025qwen2,chen2024expanding,ye2024mplug,young2024yi,team2025gemma,claude35web,awadalla2023openflamingo,li2023blip,peng2023kosmos} demonstrate strong performance in general-purpose scenarios by jointly modeling multiple modalities such as text, image and video. Motivated by their success in open-domain scenarios, researchers have extended the MLLM paradigm to the medical domain, giving rise to medical multimodal large language models (Med-MLLMs), including LLaVA-Med~\cite{li2023llava}, HuatuoGPT-Vision~\cite{chen2024huatuogpt}, HealthGPT~\cite{lin2025healthgpt}, and Lingshu~\cite{xu2025lingshu}. These models show potential for medical diagnosis and reasoning, driving progress in intelligent healthcare applications.

In fact, existing Med-MLLMs are constrained by vision-centric paradigms that focus on imaging data~\cite{nam2025multimodal,zhang2024potential}, yet medical applications frequently require integration of non-imaging modalities and domain-specific knowledge~\cite{sun2025medical}. Electrocardiograms (ECG) illustrates this limitation, as it encodes critical diagnostic information in a dual signal--image form, tightly combining temporal and morphological features~\cite{xie2020computational,ayyub2025comprehensive}. Consequently, ECG not only delineates the failure boundary of existing approaches but also highlights the fundamental rationale for developing dedicated Med-MLLMs: \textit{the structural characteristics of medical modalities must serve as the foundation of model design, rather than an afterthought for adaptation.}

Intelligent ECG interpretation relies on symbolic semantics to integrate patient history, ECG data, and clinical reports into interpretable reasoning chains. This paradigm, naturally suited to MLLMs, still faces fundamental challenges:

\noindent \textbf{Data Misalignment.} Mainstream ECG datasets such as PTB-XL~\cite{wagner2020ptb} are designed for rhythm and disease classification, with annotations at the global level and lacking step-wise supervision. As a result, these datasets mainly enable direct mapping from ECG signals to labels, making it difficult to support interpretable reasoning or cross-modal information integration, and thus limiting the shift from simple classification to deeper understanding and reasoning.

\noindent \textbf{Evaluation Deficiency.} Existing medical benchmarks primarily emphasize classification accuracy and tend to overlook the assessment of comprehensive reasoning processes~\cite{topol2019high,moor2023med}. For instance, they rarely evaluate models’ capabilities in clinical reasoning or pathology question--answering (QA) for ECG tasks.  Therefore, it is essential to develop multidimensional assessment frameworks that include reliability and scenario-based metrics for evaluating the clinical utility of ECG-specific Med-MLLMs.

\noindent \textbf{Dual Signal--Image Fusion Incapacity.} The dual nature of ECG data---comprising both raw temporal signals and waveform images---poses unique modeling challenges~\cite{prabhakararao2023multi,strodthoff2020deep}. While signals capture dynamic cardiac activity and images reflect spatial morphology, both are essential for diagnosis. Most vision-focused MLLMs, however, process only ECG images~\cite{ao2023image,nguyen2025comparing}, neglecting key temporal features in the signal, such as intervals, calibration, inter-lead phase, and beat-to-beat changes. Moreover, multi-lead synchronous acquisition results in a tightly coupled, heterogeneous structure across temporal and spatial domains~\cite{lian2022multi}. Addressing this dual complexity requires Med-MLLMs capable of consistent multimodal modeling and alignment.

We propose \textbf{Heartcare Suite}---a systematic and innovative framework for ECG, dedicated to establishing a unified and extensible ecosystem for ECG-specific Med-MLLMs:

\noindent \textbf{(i) Dataset}. We construct \textbf{Heartcare-400K}, a large-scale, fine-grained, multi-task multimodal ECG instruction dataset. It combines two sources: the public PTB-XL dataset~\cite{wagner2020ptb} with 21,799 12-lead ECG signals annotated with 179 SCP-ECG classes, and 12,170 ECG images with structured reports from top hospitals, including scanned traces, clinical conclusions, and de-identified metadata---substantially enriching modality and label diversity. To transform heterogeneous ECG data into structured annotations, we develop \textbf{HeartAgent}, a multimodal engine with a bottom-up pipeline that ensures annotation consistency and generates high-quality instruction-style QA pairs. 

\noindent \textbf{(ii) Benchmark}. We propose \textbf{Heartcare-Bench}, the first fine-grained, multidimensional evaluation framework for ECG diagnostic intelligence, designed to assess a spectrum of model capabilities ranging from feature recognition to reasoning. Built upon Heartcare-400K, Heartcare-Bench systematically covers five major task types---\textit{Closed-QA}, \textit{Open-QA}, \textit{Comparison-QA}, \textit{Report Generation}, and \textit{Signal Prediction}---spanning key diagnostic dimensions such as rhythm, waveform, and morphology. It comprises three complementary modality subsets: \textit{Signal} (S), \textit{Image} (I), and \textit{Cross-Modal} (C), enabling unified evaluation from single-modality reasoning to multi-ECG semantic alignment. With a hierarchical, multi-metric scoring system, Heartcare-Bench integrates knowledge reasoning and cross-modal understanding within a unified evaluation coordinate. 

\noindent \textbf{(iii) Model.} The dual-form characteristics of ECG introduce unique structural complexity in modeling. We propose \textbf{HeartcareGPT}, aiming to build ECG-specific Med-MLLMs. We design \textbf{Bidirectional ECG Abstract Tokenization (Beat)}, a structure-aware discrete encoding mechanism centered on vector quantization~\cite{esser2021taming,van2017neural,zeghidour2021soundstream}, which maps high-frequency continuous signals into token sequences. The design comprises three components: (i) \textit{Dual-level Vector Quantization (DVQ)}, which refines rhythm and inter-lead phase dependencies captured by the codebook to achieve high-fidelity compression; (ii) \textit{Query-guided Bidirectional Diffusion (QBD)}, which jointly models past and future contexts within the latent token space to support both signal reconstruction and prediction; and (iii) \textit{Joint Supervision Strategy}, which jointly optimizes reconstruction and prediction to maximize clinical semantic fidelity during encoding. Furthermore, we propose \textbf{Dual Stream Projection Alignment (DSPA)}, which employs dual experts to separately process ECG inputs. Through distinct preprocessing strategies and modality-specific encoders, ECG representations are transformed into embeddings compatible with Med-MLLMs. All modality embeddings are projected into a shared language space and concatenated into a unified sequence, enabling cross-modal joint reasoning for ECG under a unified autoregressive paradigm.

Experimental results demonstrate the powerful paradigm of Heartcare Suite. The main contributions of this work are as follows:

\begin{itemize}
\item \textbf{High-quality ECG Instruction Dataset.}  Heartcare-400K serves as the first comprehensive  ECG instruction dataset, which significantly enhances Med-MLLM performance across ECG-related tasks.
\item \textbf{Multidimensional ECG Benchmark.} We propose Heartcare-Bench, an evaluation framework that assesses clinical performance of ECG tasks for Med-MLLMs.
\item \textbf{Fine-grained ECG Understanding Paradigm.} We develop HeartcareGPT, the first model supporting pathology-level ECG understanding with state-of-the-art (SOTA) results.
\end{itemize}
\section{Related Work}

\noindent \textbf{Multimodal Representation Learning for ECG.} Recent advances in multimodal ECG representation learning follow three main directions. First, signal-semantic alignment. ECG-SL~\cite{yu2023ecg} and MERL~\cite{liu2024zero} align heartbeats and clinical reports via self-supervision and knowledge prompting, while HeartLang~\cite{jin2025reading} decomposes waveforms into semantic tokens for fine grained cardiac analysis. Second, cross-lead fusion. ECG-DAN~\cite{chen2025large} adopts a dual attention network to balance global cross lead interactions with local temporal dynamics, and ESI~\cite{yu2024ecg} adds a signal text contrastive learning to strengthen robustness under limited labels. Third, large language model (LLM)-driven pretraining. ECG-LM~\cite{yang2025ecg} maps ECG embeddings into a pretrained language space, and SuPreME~\cite{cai2025supreme} structures domain knowledge from clinical reports for pretraining. However, existing approaches typically address isolated aspects such as reconstruction quality, lead interaction, or semantic alignment, without forming a unified end-to-end multimodal ECG modeling framework.

\noindent \textbf{Medical Multimodal Large Language Models.} Med-MLLMs have shown strong capabilities in medical understanding and diagnostic support~\cite{chen2024huatuogpt, pan2025medvlm, jiang2024medmoemixturedomainspecificexperts,xu2025lingshu}. Med-Flamingo~\cite{moor2023med} and LLaVA-Med~\cite{li2023llava} are representative early medical multimodal models, focusing on image--text alignment and visual question answering. MedVLM~\cite{pan2025medvlm} employs multi-stage pretraining and attains SOTA performance in radiology report generation and organ localization. HealthGPT~\cite{lin2025healthgpt} unifies image understanding and generation within one framework. Domain-specific variants---LLaVA-Rad~\cite{chaves2024towards}, EyecareGPT~\cite{li2025eyecaregpt}, and SkinGPT-4~\cite{zhou2023skingpt}---enable structured reporting and multimodal reasoning across radiology, ophthalmology, and dermatology. 
However, the dual-form characteristics of ECG presents significant challenges to vision-centric models, and an effective ECG-specific Med-MLLM framework remains absent to date.
\section{Heartcare Suite: Heartcare-400K}

\subsection{Data Collection and Organization}

Existing ECG datasets suffer from limited modalities, coarse annotations, insufficient scale, and substantial heterogeneity in signal sampling, lead configuration, and preprocessing, all of which hinder the development of Med-MLLMs for intelligent ECG diagnosis. To address these gaps, we introduce \textbf{Heartcare-400K}, a large-scale multimodal ECG QA dataset with two complementary modalities: \textbf{(i) structured digital signals} and \textbf{(ii) unstructured ECG report images}. We normalize them by rendering all signals into a unified ECG visual format consistent with clinical layout. These visualized signals, together with native report images, enrich modality diversity and support comprehensive ECG understanding.

To construct Heartcare-400K, we collaborated with several major public hospitals to collect 12,170 standardized PDF-format clinical ECG reports. These reports include patient demographics, physiological parameters, physician diagnoses, and about 5 seconds of 12-lead waveform images. In addition, we systematically integrated multiple publicly available digital signal ECG datasets, with PTB-XL~\cite{wagner2020ptb} as the primary source. PTB-XL is one of the largest open ECG repositories, containing 21,799 12-lead records sampled at 500 Hz over 10 seconds, with standardized diagnostic labels and detailed patient metadata (such as gender, age and weight).

Raw ECG data typically contain only brief diagnostic text and lack the instruction-style supervision and semantic diversity needed for MLLM fine-tuning. To address this, we developed a multimodal data engine for automated extraction, cleaning, standardization, and expert review, and further augmented the dataset with QA pairs from the publicly available ECG-QA~\cite{oh2023ecg}, which supports both single-modality queries and cross-modality reasoning tasks.

Ultimately, Heartcare-400K is organized into four types of QA tasks: (i) \textit{Closed-QA} (single-ECG multiple-choice question), (ii) \textit{Open-QA} (single-ECG short-form question), (iii) \textit{Comparison-QA} (multi-ECG multiple-choice question), (iv) \textit{Report Generation} (long-form answers), and (v) \textit{Signal Prediction} (ECG generation). As Figure~\ref{fig:HeartcareGPT} shows, these tasks equip models with fine-grained ECG comprehension and clinical reasoning capabilities, making Heartcare-400K a foundational resource for developing practical and generalizable intelligent ECG diagnosis systems.

\begin{figure}[ht]
\centering
\vspace{-3mm}
\includegraphics[width=0.9\columnwidth]{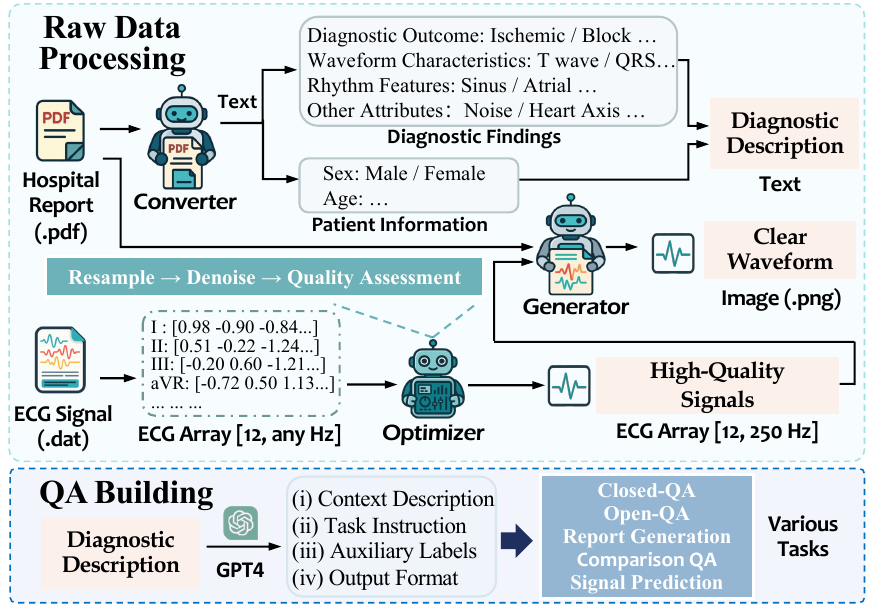}
\caption{Framework of HeartAgent for QA generation.}
\label{fig:engine}
\vspace{-3mm}
\end{figure}

\subsection{Multimodal Data Engine}

To efficiently construct Heartcare-400K, we design \textbf{HeartAgent} (as shown in Figure~\ref{fig:engine}), an automated multimodal data engine with two key stages: \textit{raw data processing} and \textit{data building}. Details of each module are provided in the Appendix~\ref{app:engine}.

\noindent \textbf{Raw Data Processing.}This stage integrates three core components to transform heterogeneous ECG sources into standardized inputs: \textbf{(i) Multimodal Feature Converter.} Extracts patient information and clinical descriptions from hospital PDFs, maps diagnostic text to structured labels, and prepares both text and image data for subsequent processing. \textbf{(ii) Noise Filtering and Quality Optimizer.} Unifies signal formats and applies denoising, resampling, and quality assessment to produce high-quality, standardized ECG signals. \textbf{(iii) Diversified Image Generator.} Renders clean waveform images from processed digital signals and generates de-identified tracings from hospital reports, ensuring multi-modal data alignment.

\noindent \textbf{Data Building.} This stage is handled by the \textbf{Multi-task QA Builder}, which constructs instruction-style QA samples using GPT-4~\cite{gpt4v}. Each sample contains structured context, clear task instructions, auxiliary labels, and standardized answers, supporting four diverse ECG understanding tasks and a \textit{Signal Prediction} task critical for comprehensive model training.

These components collectively ensure Heartcare-400K achieves high-quality, diverse, and task-relevant multimodal annotations.
\section{Heartcare Suite: Heartcare-Bench}

To systematically evaluate the capabilities of Med-MLLMs across unified ECG understanding and prediction tasks, we propose Heartcare-Bench---a multidimensional benchmark encompassing five major task types: \textit{Closed-QA}, \textit{Open-QA}, \textit{Comparison-QA}, \textit{Report Generation}, and \textit{Signal Prediction}. Heartcare-Bench is structured into three complementary subsets based on input modality: \textbf{(i) Heartcare-Bench\textsuperscript{S}} (signal-based), \textbf{(ii) Heartcare-Bench\textsuperscript{I}} (image-based), and \textbf{(iii) Heartcare-Bench\textsuperscript{C}} (cross-modal comparison). In addition, Heartcare-Bench employs strict patient-level data partitioning with thorough cross-split duplicate inspection; full details are provided in Appendix~\ref{app:bench}.

\noindent \textbf{Heartcare-Bench\textsuperscript{S/I}: Single-Modality Subsets.} The signal and image subsets share a unified framework to evaluate four core clinical dimensions: \textit{Diagnosis}, \textit{Waveform analysis}, \textit{Rhythm interpretation} and \textit{Miscellaneous Features}. To broaden task diversity and clinical relevance, we further integrate QA pairs from ECG-QA~\cite{oh2023ecg}, primarily those derived from PTB-XL, enriching \textit{Closed-QA} and \textit{Open-QA} settings and enabling categorization by these four dimensions. The \textit{Miscellaneous (Misc.)} category further assesses ECG attributes such as noise, infarction stage, ectopic beats, and cardiac axis. In addition to QA tasks, the single-modality subsets also include \textit{Report Generation} and \textit{Signal Prediction} tasks. Detailed formats and examples are provided in Appendix~\ref{app:bench}.

\noindent \textbf{Heartcare-Bench\textsuperscript{C}: Cross-Modality Comparison Subset.} Beyond single-modality tasks, this subset evaluates a model’s ability to reason over multiple ECGs through comparison QA tasks adapted from ECG-QA~\cite{oh2023ecg}. It is centered around three modality comparison dimensions: (i) \textit{signal--signal (S--S)}, (ii) \textit{image--image (I--I)}, and (iii) \textit{signal--image (S--I)}. Each configuration is further categorized into two clinically relevant subtypes: consecutive (ECGs from the same patient, assessing temporal consistency) and irrelevant (ECGs from different patients, evaluating differential diagnostic reasoning).

Overall, Heartcare-Bench adopts a multidimensional evaluation protocol covering semantic consistency, linguistic fluency, clinical accuracy, and waveform prediction quality. Full scoring rules and evaluation details are provided in Appendix~\ref{app:bench}.
\section{Methodology}

\begin{figure*}[ht]
\centering
\includegraphics[width=0.9\textwidth]{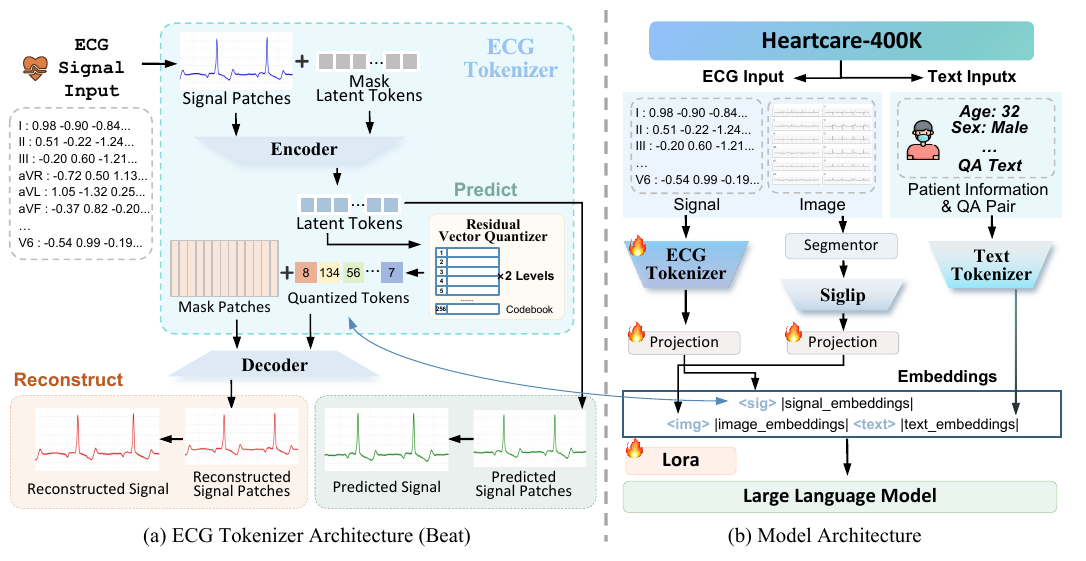}
\vspace{-3mm}
\caption{Architecture of HeartcareGPT. (a) The dual-form ECG inputs are routed and encoded with modality-specific expert projections aligned to the LLM backbone. (b) The unified autoregressive architecture efficiently supports interleaved and joint modeling of ECG multimodal inputs.}
\label{fig:method}
\vspace{-5mm}
\end{figure*}

\subsection{Bidirectional ECG Abstract Tokenization }
\label{sec:beat}

We present a structure-aware discrete encoding mechanism centered on vector quantization---\textbf{B}idirectional \textbf{E}CG \textbf{A}bstract \textbf{T}okenization \textbf{(Beat)}. The architecture of Beat is depicted in Figure~\ref{fig:method} (a).

\noindent \textbf{Forward Diffusion Process.} Given an acquired ECG signal, we denoise and resample it to obtain a representative segment $\mathbf{S}\in\mathbb{R}^{T\times C}$ (sequence length $T$, $C$ leads). We then partition $\mathbf{S}$ into contiguous, non-overlapping temporal patches and map them with a projection layer $\Psi$ to yield the corresponding patch embeddings:

\vspace{-4mm}
\begin{equation}
\mathbf{e} = \Psi(\textrm{Patchify}_f(\mathbf{S})\mid\theta_\Psi) \in \mathbb{R}^{t \times c},\, t=\frac{T}{f},c=Cf,
\end{equation}
\vspace{-4mm}

where $f$ is the patch frame size, $t$ the number of patches, and $\theta_{\Psi}$ the projection parameters. To inject high-level semantic control, we append $m$ learnable queries $\mathbf{q}\in\mathbb{R}^{m\times c}$ to the patch embeddings $\mathbf{e}$, forming the model input $\mathcal{H}_{\text{in}}=[\mathbf{e}\|\mathbf{q}]$. A Transformer encoder then performs forward diffusion, yielding compressed contextual representations:

\vspace{-4mm}
\begin{equation}
\mathcal{H}_\text{latent}^q = \left(\textrm{TransformerEnc}(\mathcal{H}_\text{in})\right)_{[t:t+m]} \in \mathbb{R}^{m \times c}.
\label{eq:forward}
\end{equation}
\vspace{-4mm}

\noindent \textbf{Dual-level Vector Quantization.} To compress ECG signals while preserving rhythmic structure and salient pathological cues, we adopt dual-level VQ with a core codebook $\mathcal{C}_1$ and a residual codebook $\mathcal{C}_2$. For each query vector $\mathbf{h}_q^i=(\mathcal{H}^{q}_{\text{latent}})_{[i,:]}\in\mathbb{R}^{c}$, we first apply the core quantization:

\vspace{-4mm}
\begin{equation}
\hat{\mathbf{h}}_{\langle q,1\rangle}^{i} = \textrm{Quant}_{\mathcal{C}_1}(\mathbf{h}_q^i) = \arg\min_{\mathbf{c} \in \mathcal{C}_1} \|\mathbf{h}_q^i - \mathbf{c}\|_2.
\end{equation}
\vspace{-4mm}

The residual is then quantized by the codebook $\mathcal{C}_2$:

\vspace{-4mm}
\begin{equation}
\hat{\mathbf{h}}_{\langle q,2\rangle}^{i} = \textrm{Quant}_{\mathcal{C}_2}(\mathbf{h}_q^i - \hat{\mathbf{h}}_{\langle q,1\rangle}^{i}) = \arg\min_{\mathbf{c} \in \mathcal{C}_2} \|\mathbf{h}_q^i - \hat{\mathbf{h}}_{\langle q,1\rangle}^{i} - \mathbf{c}\|_2.
\end{equation}
\vspace{-4mm}

Finally, we quantize the forward-diffused representation to $\hat{\mathbf{h}}_{\text{latent}}^{i}=\hat{\mathbf{h}}_{\langle q,1\rangle}^{i}+\hat{\mathbf{h}}_{\langle q,2\rangle}^{i}$, yielding a high-fidelity mapping from continuous latents to a discrete code space and significantly improving reconstruction quality.

\begin{table*}[ht]
\centering
\renewcommand{\arraystretch}{1.2}
\caption{Performance comparison between HeartcareGPT and other baselines on \textit{Closed-QA} tasks from Heartcare-Bench\textsuperscript{S} and Heartcare-Bench\textsuperscript{I}, evaluated by accuracy. We use \textbf{bold} text to indicate the best results and \underline{underline} to indicate the second-best results.}
\label{tab:close}
\begin{adjustbox}{width=0.7\textwidth,center}
\begin{tabular}{lcccccccccc}
\toprule
\multirow{2}{*}{\textbf{Model}} 
& \multicolumn{4}{c}{\textbf{Heartcare-Bench\textsuperscript{S}}} 
& \multicolumn{4}{c}{\textbf{Heartcare-Bench\textsuperscript{I}}} 
& \multirow{2}{*}{\textbf{Avg.}} \\
\cmidrule(lr){2-5} \cmidrule(lr){6-9}
& \textbf{Diagnosis} & \textbf{Waveform} & \textbf{Rhythm} & \textbf{Misc.} 
& \textbf{Diagnosis} & \textbf{Waveform} & \textbf{Rhythm} & \textbf{Misc.} 
& \\
\hline
\rowcolor{gray!10} 
\multicolumn{10}{c}{\textbf{\textit{Generalist Models}}} \\
\hline
LLaVA-1.5-7B~\cite{liu2023visual} & 19.26 & 27.03 & 22.19 & 28.96 & 15.15 & 12.40 & 18.20 & 25.47 & 21.08\\
Qwen2.5-VL-7B~\cite{bai2025qwen2} & 28.29 & 35.36 & 31.70 & 35.31 & 35.84 & 31.06 & 22.35 & 41.41 & 32.67\\
InternVL-2.5-8B~\cite{chen2024expanding} & 29.87 & 34.18 & 31.25 & 40.75 & 25.86 & 17.89 & 38.23 & 40.05 & 32.26\\
Yi-VL-6B~\cite{young2024yi} & 49.19 & 38.90 & 31.06 & 39.65 & 59.55 & 44.46 & 17.06 & 39.85 & 39.97\\
MiniCPM-V2.6-8B~\cite{hu2024minicpm} & 32.37 & 31.34 & 23.40 & 32.23 & 23.71 & 33.59 & 10.05 & 30.63 & 27.17\\
Gemma-3-4B~\cite{team2025gemma} & 23.84 & 24.36 & 18.30 & 21.47 & 38.73 & 38.09 & 20.63 & 19.32 & 25.59\\
Claude-3.5~\cite{claude35web} & 37.37 & 20.73 & 15.96 & 35.16 & 30.58 & 37.76 & 37.83 & 34.45 & 31.23\\
GPT5~\cite{gpt5} & 29.87 & 15.23 & 19.57 & 25.23 & 41.42 & 34.14 & 33.73 & 26.23 & 28.30\\
\hline
\rowcolor{gray!10}
\multicolumn{10}{c}{\textbf{\textit{Medical Models}}} \\
\hline
Lingshu-7B~\cite{xu2025lingshu} & 29.56 & 33.44 & 40.00 & 20.75 & 47.96 & 37.98 & 19.31 & 30.28 & 32.41\\
LLaVA-Med-7B~\cite{li2023llava} & 27.79 & 25.86 & 24.04 & 33.69 & 25.11 & 25.96 & 26.17 & 33.36 & 27.75\\
MedVLM-R1-2B~\cite{pan2025medvlm} & 16.27 & 15.42 & 15.32 & 36.32 & 14.08 & 19.43 & 8.33 & 35.56 & 20.09\\
HealthGPT-M3-3.8B~\cite{lin2025healthgpt} & 21.63 & 10.87 & 26.58 & 20.23 & 15.35 & 12.41 & 16.06 & 23.97 & 18.39\\
\rowcolor{lightblue}
\textbf{HeartcareGPT-3.8B} & \underline{81.95} & \underline{95.94} & \underline{82.79} & \textbf{79.84} & \underline{87.85} & \textbf{92.21} & \underline{79.25} & \underline{67.80} & \underline{83.33} \\
\rowcolor{lightblue}
\textbf{HeartcareGPT-7B} & \textbf{86.16} & \textbf{98.62} & \textbf{93.20} & \underline{75.15} & \textbf{62.71} & \underline{87.82} & \textbf{85.13} & \textbf{78.59} & \textbf{83.42}\\
\bottomrule
\end{tabular}
\end{adjustbox}
\vspace{-3mm}
\end{table*}

\noindent \textbf{Query-guided Bidirectional Diffusion.} Forward diffusion (Eq.~\ref{eq:forward}) compresses the ECG into a dense discrete space but offers no guarantees of completeness or invertibility. We therefore adopt an autoencoding view and introduce a reverse diffusion anchored at the quantized query $\mathcal{H}_{\text{latent}}^{q}$, casting compression and reconstruction as a joint constraint: the forward path enforces compact coding, the reverse path restores details, yielding a closed-loop, bidirectional token refinement between latent and signal spaces.

During reverse diffusion, the original input $\mathbf{e}$ is mapped to its forward-diffusion slot $\mathcal{H}^q_{\text{origin}}$ and masked by $\mathcal{M}$ to prevent leakage, so only the query vectors $\mathcal{H}^q_{\text{latent}}$ retain ECG signal information for reconstruction:

\vspace{-4mm}
\begin{equation}
\mathcal{H}_\text{out}^\text{recon} = \left(\text{TransformerDec}(\mathcal{H}_\text{latent}^q; \mathcal{M})\right)_{[0:t]}.
\end{equation}
\vspace{-4mm}

\noindent \textbf{Joint Supervision Strategy.} To fully exploit the bidirectional diffusion capacity, Beat employs a multi-objective loss that jointly optimizes reconstruction and compression, integrating three objectives. The reconstruction and prediction losses are defined as:

\vspace{-4mm}
\begin{equation}
\mathcal{L}_\text{recon} = \|\mathcal{H}_\text{out}^\text{recon} - \mathbf{e}\|_2, \qquad \mathcal{L}_\text{pred} = \|\mathcal{H}_\text{latent}^q - \mathbf{e}_\text{pred}\|_2.
\end{equation}
\vspace{-4mm}

where $\mathbf{e}_{\text{pred}}$ denotes the ECG segment used for prediction. The vector quantization loss is defined as:

\vspace{-4mm}
\begin{equation}
\mathcal{L}_{\text{VQ}} = \sum_{i,j} \|\text{sg}[\mathbf{h}_i^i] - \hat{\mathbf{h}}_{\langle q,i\rangle}^i\|_2^2 + \beta \|\mathbf{h}_j^i - \text{sg}[\hat{\mathbf{h}}_{\langle q,j\rangle}^i]\|_2^2,
\end{equation}
\vspace{-4mm}

where $\text{sg}[\cdot]$ denotes the stop-gradient operation, and $\mathbf{h}_j^i$ refers to the feature vector before quantization at the $j$-th level. The overall training objective is given by $\mathcal{L}_{\text{total}} = \lambda_1 \mathcal{L}_{\text{recon}} + \lambda_2 \mathcal{L}_{\text{pred}} + \lambda_3 \mathcal{L}_{\text{VQ}}$.

\noindent \textbf{Tokenization.} During inference, we discard the decoder and prediction heads, retaining only the encoder and quantizer. Given an ECG segment $\mathbf{S}$ of length $T$, Beat applies temporal normalization: if $T>t$, split into $\lceil T/t \rceil$ slices from right to left; if $T<t$, left-pad to $t$. For recordings with fewer than 12 leads, pad missing channels with zeros. Each segment is then encoded into discrete tokens:

\vspace{-4mm}
\begin{equation}
\mathcal{S} = \text{Beat}(\mathbf{S}) = \{c^1_1,\ldots,c^n_1\|c^1_2, \ldots,c^n_2\},\, c_i^j \in \{\mathcal{C}_1, \mathcal{C}_2\}.
\label{eq:Beat}
\end{equation}
\vspace{-4mm}

These discrete tokens can be directly used as multimodal extensions of LLM vocabulary, enabling unified semantic modeling and cross-modal reasoning between ECG signals and texts.

For autoregressive forecasting, Beat predicts the next $t'$ segment conditioned on the previous context $(T+t')$, iteratively generating arbitrary-length ECG sequences.

\subsection{HeartcareGPT}

\begin{table*}[ht]
\centering
\renewcommand{\arraystretch}{1.2}
\caption{Performance comparison between HeartcareGPT and baselines on \textit{Open-QA} tasks from Heartcare-Bench\textsuperscript{S} and Heartcare-Bench\textsuperscript{I}.}
\label{tab:open}
\begin{adjustbox}{width=0.98\textwidth,center}
\begin{tabular}{lcccccccccccccccc}
\toprule
\multirow{3}{*}{\textbf{Model}} & 
\multicolumn{8}{c}{\textbf{Heartcare-Bench\textsuperscript{S}}} & 
\multicolumn{8}{c}{\textbf{Heartcare-Bench\textsuperscript{I}}} \\
\cmidrule(lr){2-9}
\cmidrule(lr){10-17}
& \multicolumn{2}{c}{\textbf{Diagnosis}} 
& \multicolumn{2}{c}{\textbf{Waveform}} 
& \multicolumn{2}{c}{\textbf{Rhythm}} 
& \multicolumn{2}{c}{\textbf{Misc.}} 
& \multicolumn{2}{c}{\textbf{Diagnosis}} 
& \multicolumn{2}{c}{\textbf{Waveform}} 
& \multicolumn{2}{c}{\textbf{Rhythm}} 
& \multicolumn{2}{c}{\textbf{Misc.}} \\
\cmidrule(lr){2-3} 
\cmidrule(lr){4-5} 
\cmidrule(lr){6-7} 
\cmidrule(lr){8-9} 
\cmidrule(lr){10-11} 
\cmidrule(lr){12-13} 
\cmidrule(lr){14-15} 
\cmidrule(lr){16-17}
& \textbf{F1-Bio} & \textbf{Rouge-L} 
& \textbf{F1-Bio} & \textbf{Rouge-L} 
& \textbf{F1-Bio} & \textbf{Rouge-L} 
& \textbf{F1-Bio} & \textbf{Rouge-L} 
& \textbf{F1-Bio} & \textbf{Rouge-L} 
& \textbf{F1-Bio} & \textbf{Rouge-L} 
& \textbf{F1-Bio} & \textbf{Rouge-L} 
& \textbf{F1-Bio} & \textbf{Rouge-L} \\
\hline
\rowcolor{gray!10}\multicolumn{17}{c}{\textbf{\textit{Generalist Models}}}\\
\hline
LLaVA-1.5-7B~\cite{liu2023visual} & 18.52 & 11.57 & 25.34 & 13.93 & 20.84 & 12.45 & 27.53 & 14.82 & 14.82 & 10.25 & 11.24 & 9.67 & 17.53 & 11.38 & 24.21 & 13.76\\
Qwen2.5-VL-7B~\cite{bai2025qwen2} & 27.63 & 12.84 & 34.15 & 15.72 & 30.82 & 13.96 & 34.27 & 16.38 & 35.24 & 14.93 & 30.35 & 13.27 & 21.84 & 12.18 & 40.13 & 17.24\\
InternVL-2.5-8B~\cite{chen2024expanding} & 29.27 & 13.29 & 33.52 & 14.86 & 30.64 & 13.72 & 39.82 & 16.95 & 25.37 & 11.82 & 17.26 & 10.45 & 37.53 & 15.63 & 39.27 & 16.84\\
Yi-VL-6B~\cite{young2024yi} & 48.53 & 15.38 & 38.27 & 14.72 & 30.42 & 13.85 & 38.96 & 15.63 & 58.92 & 17.82 & 43.85 & 16.27 & 16.43 & 10.84 & 38.74 & 15.49\\
MiniCPM-V2.6-8B~\cite{hu2024minicpm} & 31.84 & 12.93 & 30.73 & 12.84 & 22.86 & 11.75 & 31.52 & 13.92 & 23.17 & 11.28 & 32.94 & 13.46 & 9.53 & 8.67 & 29.83 & 13.27\\
Gemma-3-4B~\cite{team2025gemma} & 12.90 & 5.49 & 11.57 & 6.95 & 13.85 & 5.87 & 23.01 & 6.08 & 14.13 & 6.65 & 14.74 & 9.90 & 15.12 & 5.95 & 21.02 & 8.98\\
Claude-3.5~\cite{claude35web} & 21.43 & 5.04 & 14.81 & 4.36 & 23.01 & 6.75 & 17.23 & 7.91 & 21.43 & 5.39 & 27.69 & 5.02 & 17.78 & 4.13 & 19.14 & 10.24\\
GPT5~\cite{gpt5} & 23.00 & 7.25 & 21.42 & 13.93 & 37.39 & 14.71 & 24.83 & 9.57 & 40.82 & 15.93 & 33.52 & 12.84 & 33.17 & 12.53 & 25.64 & 9.82\\
\hline
\rowcolor{gray!10}\multicolumn{17}{c}{\textbf{\textit{Medical Models}}}\\
\hline
Lingshu-7B~\cite{xu2025lingshu} & 28.94 & 12.37 & 32.84 & 14.26 & 39.27 & 15.84 & 20.17 & 10.53 & 47.36 & 16.28 & 37.35 & 15.12 & 18.74 & 10.87 & 29.63 & 13.15\\
LLaVA-Med-7B~\cite{li2023llava} & 27.15 & 12.84 & 25.27 & 11.93 & 23.48 & 11.27 & 33.02 & 14.38 & 24.53 & 11.29 & 25.37 & 12.15 & 25.53 & 11.84 & 32.74 & 14.26\\
MedVLM-R1-2B~\cite{pan2025medvlm} & 15.64 & 8.73 & 14.83 & 8.25 & 14.73 & 8.16 & 35.64 & 13.52 & 13.52 & 7.84 & 18.84 & 9.27 & 16.83 & 6.45 & 34.92 & 13.17\\
HealthGPT-M3-3.8B~\cite{lin2025healthgpt} & 21.07 & 10.82 & 10.27 & 8.93 & 25.94 & 11.76 & 19.63 & 10.24 & 14.83 & 9.67 & 11.84 & 8.75 & 15.43 & 9.82 & 23.37 & 11.39\\
\rowcolor{lightblue}
\textbf{HeartcareGPT-3.8B} & \underline{68.53} & \underline{32.27} & \underline{72.74} & \underline{35.38} & \textbf{78.63} & \textbf{36.42} & \underline{65.84} & \underline{30.97} & \underline{63.17} & \underline{29.83} & \underline{70.92} & \textbf{33.57} & \textbf{75.36} & \textbf{34.69} & \underline{61.53} & \underline{28.24}\\
\rowcolor{lightblue}
\textbf{HeartcareGPT-7B} & \textbf{72.94} & \textbf{36.85} & \textbf{78.26} & \textbf{39.64} & \underline{73.62} & \underline{35.27} & \textbf{68.37} & \textbf{34.18} & \textbf{67.43} & \textbf{33.52} & \textbf{76.73} & \textbf{37.84} & \underline{69.84} & \underline{32.47} & \textbf{64.17} & \textbf{31.73}\\
\bottomrule
\end{tabular}
\end{adjustbox}
\vspace{-3mm}
\end{table*}

As Figure~\ref{fig:method} (b) shows, we present \textbf{HeartcareGPT}, which maps ECG signals, ECG images, and text into a unified discrete space, enabling ECG reasoning with a single autoregressive architecture.

\noindent \textbf{Unified ECG Tokens.} Specifically, for a multi-lead ECG $S\in\mathbb{R}^{T\times C}$, Beat (Eq.~\ref{eq:Beat}) encodes it into a token sequence $\mathcal{S}$ capturing elementary ECG morphology. For ECG images, we partition and rearrange them into 12 lead-wise maps to decouple lead semantics, then apply SigLip~\cite{zhai2023sigmoidlosslanguageimage} to obtain visual tokens $\mathcal{V}$ that embed per-lead waveform and spatial layout. Meanwhile, clinical notes, basic physiological data, and diagnostic instructions are tokenized by the native tokenizer of the LLM into textual features $\mathcal{F}_{\mathcal{T}}$, providing a semantic anchor for non-text modalities and enabling fine-grained alignment and joint modeling in the shared token space.

\noindent \textbf{Dual Stream Projection Alignment.} To avoid parameter interference among modalities in shallow layers, we adopt a decoupled design, mapping sequential feature $\mathcal{F}_{\mathcal{S}}$ and visual feature $\mathcal{F}_{\mathcal{V}}$ into the model’s embedding space through dedicated expert projections $\Psi_{\text{sig}}$ and $\Psi_{\text{img}}$:

\vspace{-4mm}
\begin{equation}
\mathcal{F}_\mathcal{S} = \Psi_{\text{sig}}(\mathcal{S}\mid\theta_{\Psi_\text{sig}}), \quad
\mathcal{F}_\mathcal{V} = \Psi_{\text{img}}(\mathcal{V}\mid\theta_{\Psi_\text{img}}),
\end{equation}
\vspace{-4mm}

so that they lie in the same representation space. Either projection layer is implemented as a lightweight MLP block. Subsequently, at the input side we directly concatenate $\mathcal{F}_\mathcal{S}$, $\mathcal{F}_\mathcal{V}$ and $\mathcal{F}_\mathcal{T}$ into a single long sequence:

\vspace{-4mm}
\begin{equation}
\langle \texttt{BOS}\rangle\langle \texttt{sig}\rangle\{\mathcal{F}_\mathcal{S}\}\langle \texttt{img}\rangle\{\mathcal{F}_\mathcal{V}\}\langle \texttt{text}\rangle\{\mathcal{F}_\mathcal{T}\}\langle \texttt{EOS}\rangle,
\end{equation}
\vspace{-4mm}

where $\langle \texttt{sig}\rangle$ and $\langle \texttt{img}\rangle$ are specially introduced conditional tokens that can be expanded as needed at the implementation level to accommodate multiple signal segments and multiple images. With this design, the routing and composition of multimodal inputs no longer rely on additional architectural components, but are uniformly reduced to simple template filling of instruction prompts.

\noindent \textbf{Training Strategy.} To enable stable coexistence of tokens from heterogeneous modalities and distributions within the base model, HeartcareGPT adopts an overall training pipeline with clearly separated roles: \textbf{(i) Multimodal warm-up}, where we only train the signal and image projection layers using a captioning task to preliminarily align multimodal features with textual features; \textbf{(ii) Joint fine-tuning}, where, once all modalities have been embedded into a compatible space, we unfreeze the weights of the LLM and optimize it together with the modality projection layers in an end-to-end manner on instruction tasks.

\noindent \textbf{Autoregressive Multimodal Generation.} The unified training objective of the model is to generate the corresponding textual output $\mathcal{R} = [r_1, r_2, \ldots, r_{N_r}]$ conditioned on the above multimodal input $\mathcal{U}=\{\mathcal{F}_\mathcal{S},\mathcal{F}_\mathcal{V},\mathcal{F}_\mathcal{T}\}$, by maximizing the following probability distribution:

\vspace{-4mm}
\begin{equation}
P_\theta(\mathcal{R} \mid \mathcal{U}) = \prod_{j=1}^{N_r} P_\theta(r_j \mid \{\mathcal{U}\| r_{<j}\}).
\end{equation}
\vspace{-4mm}

Here, $\theta$ denotes the parameters of $\mathcal{M}_{\text{llm}}$. The above optimization equips the model with strong capabilities in ECG diagnosis and cross-modal knowledge complementation, yielding a general paradigm for ECG specific Med-MLLMs.
\section{Experiments}
\label{sec:Experiments}

\subsection{Data and Experimental Setup}

\noindent \textbf{Data Details.} We systematically evaluate our model with baseline models on the proposed Heartcare-Bench\textsuperscript{S}, Heartcare-Bench\textsuperscript{I} and Heartcare-Bench\textsuperscript{C}, providing a comprehensive assessment of generalization and diagnostic accuracy. 
For baseline models that cannot accept digital signal input directly, we convert signals into images. 
More details are provided in Appendix~\ref{app:dataset}.

\noindent \textbf{Model Details.} HeartcareGPT employs SigLip~\cite{zhai2023sigmoidlosslanguageimage} and the proposed Beat as dual-form feature encoders and adopts a three-stage training paradigm: (i) training Beat to extract high-fidelity ECG embeddings, (ii) warming up the visual and signal projectors to stabilize feature alignment, and (iii) performing joint instruction fine-tuning on Heartcare-400K for end-to-end modeling. Detailed hyperparameter settings and model configurations are provided in Appendix~\ref{app:train}.

\noindent \textbf{Baseline.} We conduct a zero-shot evaluation on 12 representative LLMs, including eight open-world LLMs (e.g., LLaVA-v1.5~\cite{li2023llava}, Qwen2.5-VL~\cite{bai2025qwen2}, InternVL2.5~\cite{chen2024expanding}, Yi-VL~\cite{young2024yi}, MiniCPM-V2.6~\cite{hu2024minicpm}, gemma-3~\cite{team2025gemma}, Claude-3.5~\cite{claude35web}, GPT5~\cite{gpt5} and four Med-MLLMs (e.g., Lingshu-7B~\cite{xu2025lingshu}, LLaVA-Med~\cite{li2023llava}, MedVLM-R1~\cite{pan2025medvlm}, HealthGPT~\cite{lin2025healthgpt}). \textit{Signal prediction} tasks are not included in the evaluation when baseline models fail to respond to prediction instructions correctly. More details refer to Appendix~\ref{app:model}.

\subsection{Main Results}
\label{sec:main_results}

\noindent \textbf{Closed-QA.} As shown in Table~\ref{tab:close}, HeartcareGPT-7B achieves SOTA performance on \textit{Closed-QA} with an average accuracy of 83.42\%, while HeartcareGPT-3.8B achieves 83.33\%, surpassing other models across all subtasks by a large margin. We attribute this to the ECG-aware tokenization and instruction tuning framework, which enables precise alignment between temporal signal patterns and clinically grounded language reasoning.

\noindent \textbf{Open-QA.} Table~\ref{tab:open} reports the results on \textit{Open-QA} tasks, evaluated using \textit{BioBERTScore-F1 (F1-Bio})~\cite{ramshaw1995text} and \textit{ROUGE-L}~\cite{lin2004rouge}. Our HeartcareGPT series achieve the highest overall performance across four subtasks, demonstrating strong capability in generating clinically relevant, semantically consistent answers grounded in ECG signals.

\noindent \textbf{Comparison-QA.} The \textit{Comparison-QA} task assesses cross-modal reasoning by requiring analysis and contrast of two ECG inputs---either signals, images, or signal--image pairs---in Heartcare-Bench\textsuperscript{C}. As shown in Table~\ref{tab:comparison}, HeartcareGPT excels in this challenging setting, with both 3.8B and 7B models achieving SOTA performance. This success underscores the effectiveness of our multimodal fusion design, which enables a unified and nuanced understanding across heterogeneous ECG representations.

\begin{table}[ht]
\centering
\renewcommand{\arraystretch}{1.2}
\caption{Performance comparison between HeartcareGPT and other baselines on \textit{Comparison-QA} tasks from Heartcare-Bench\textsuperscript{C}. \textit{Cons.} = Consecutive ECGs from the same patient; \textit{Irr.} = Irrelevant ECGs from different patients. Yi-VL-6B and HealthGPT-M3-3.8B do not support multi-image input.}
\label{tab:comparison}
\begin{adjustbox}{width=0.9\columnwidth,center}
\begin{tabular}{lccccccc}
\toprule
\multirow{2}{*}{\textbf{Model}} 
& \multicolumn{2}{c}{\textbf{S--S}} 
& \multicolumn{2}{c}{\textbf{I--I}} 
& \multicolumn{2}{c}{\textbf{S--I}} 
& \multirow{2}{*}{\textbf{Avg.}} \\

\cmidrule(lr){2-3} \cmidrule(lr){4-5} \cmidrule(lr){6-7}
& \textbf{Cons.} & \textbf{Irr.} 
& \textbf{Cons.} & \textbf{Irr.} 
& \textbf{Cons.} & \textbf{Irr.} & \\

\hline
\rowcolor{gray!10} \multicolumn{8}{c}{\textbf{\textit{Generalist Models}}} \\
\hline
LLaVA-1.5-7B~\cite{liu2023visual} & 62.70 & 56.24 & 66.17 & 63.51 & 47.94 & 45.65 & 57.04\\
Qwen2.5-VL-7B~\cite{bai2025qwen2} & 50.64 & 47.26 & 49.36 & 50.68 & 50.21 & 51.28 & 49.91\\
InternVL-2.5-8B~\cite{chen2024expanding} & 47.83 & 45.13 & 46.18 & 43.33 & 49.40 & 43.24 & 45.85\\
MiniCPM-V2.6-8B~\cite{hu2024minicpm} & 53.22 & 48.13 & 51.87 & 45.50 & 49.85 & 45.05 & 48.97\\
Gemma-3-4B~\cite{team2025gemma} & 49.18 & 38.53 & 49.63 & 35.44 & 49.55 & 35.59 & 42.99\\
Claude-3.5~\cite{claude35web} & 46.17 & 43.96 & 39.15 & 56.78 & 24.47 & 54.82 & 44.23\\
GPT5~\cite{gpt5} & 50.97 & 49.18 & 50.82 & 49.40 & 50.00 & 47.90 & 49.71\\
\hline
\rowcolor{gray!10} \multicolumn{8}{c}{\textbf{\textit{Medical Models}}} \\
\hline
Lingshu-7B~\cite{xu2025lingshu} & 58.79 & 45.01 & 62.13 & 59.46 & 45.11 & 43.12 & 52.27\\
LLaVA-Med-7B~\cite{li2023llava} & 35.46 & 34.82 & 32.84 & 32.65 & 43.26 & 34.14 & 35.53\\
MedVLM-R1-2B~\cite{pan2025medvlm} & 58.44 & 39.33 & 53.69 & 40.34 & 49.57 & 37.95 & 46.55\\
\rowcolor{lightblue}
\textbf{HeartcareGPT-3.8B} & \underline{66.40} & \underline{67.47} & \underline{69.88} & \underline{75.19} & \underline{78.71} & \underline{78.83} & \underline{72.74}\\
\rowcolor{lightblue}
\textbf{HeartcareGPT-7B} & \textbf{74.01} & \textbf{75.87} & \textbf{74.80} & \textbf{79.09} & \textbf{80.05} & \textbf{79.55} & \textbf{77.23}\\
\bottomrule
\end{tabular}
\end{adjustbox}
\vspace{-3mm}
\end{table}

\noindent \textbf{Report Generation.} Table~\ref{tab:report} presents HeartcareGPT’s \textit{Report Generation} results on Heartcare-Bench\textsuperscript{S} and Heartcare-Bench\textsuperscript{I}, evaluated by \textit{Score\textsuperscript{GPT}}, \textit{F1-Rad}~\cite{jain2021radgraph}, and \textit{ROUGE-L}~\cite{lin2004rouge}. HeartcareGPT achieves top scores on most metrics, outperforming all baseline models. While Qwen2.5-VL-7B scores higher on some metrics related to report structure and expression, HeartcareGPT excels in clinical content accuracy and reliability. Since GPT-based evaluation considers formatting and completeness, score differences often reflect output style rather than clinical accuracy. Full criteria and model comparisons are detailed in Table~\ref{tab:eval_criteria}.

\begin{table}[ht]
\centering
\renewcommand{\arraystretch}{1.2}
\caption{Performance comparison between HeartcareGPT and other baseline methods on \textit{Report Generation} tasks from Heartcare-Bench\textsuperscript{S} and Heartcare-Bench\textsuperscript{I}.}
\label{tab:report}
\begin{adjustbox}{width=0.9\columnwidth,center}
\begin{tabular}{lcccccc}
\toprule
\multirow{2}{*}{\textbf{Model}} & \multicolumn{3}{c}{\textbf{Heartcare-Bench\textsuperscript{S}}} & \multicolumn{3}{c}{\textbf{Heartcare-Bench\textsuperscript{I}}}\\
\cmidrule(lr){2-4} \cmidrule(lr){5-7}
& \textbf{Score\textsuperscript{GPT}} & \textbf{F1-Rad} & \textbf{Rouge-L} 
& \textbf{Score\textsuperscript{GPT}} & \textbf{F1-Rad} & \textbf{Rouge-L} \\
\hline
\rowcolor{gray!10} \multicolumn{7}{c}{\textbf{\textit{Generalist Models}}} \\
\hline
LLaVA-1.5-7B~\cite{liu2023visual} & 55.69 & 5.22 & 18.30 & 37.40 & 12.90 & 20.75\\
Qwen2.5-VL-7B~\cite{bai2025qwen2} & \underline{64.60} & 5.64 & 9.00 & \underline{67.40} & 7.67 & 13.56\\
InternVL-2.5-8B~\cite{chen2024expanding} & 42.07 & 5.50 & 9.13 & 33.69 & 7.73 & 11.02\\
Yi-VL-6B~\cite{young2024yi} & 26.58 & 3.75 & 13.05 & 21.74 & 6.73 & 15.18\\
MiniCPM-V2.6-8B~\cite{hu2024minicpm} & 34.54 & 5.74 & 10.66 & 49.60 & 7.48 & 12.63\\
Gemma-3-4B~\cite{team2025gemma} & 64.57 & 4.38 & 7.57 & 57.10 & 7.00 & 9.07\\
Claude-3.5~\cite{claude35web} & 63.11 & 5.03 & 12.33 & 63.29 & 7.63 & 14.02\\
GPT5~\cite{gpt5} & 62.73 & 4.85 & 8.86 & 78.80 & 6.29 & 9.55\\
\hline
\rowcolor{gray!10} \multicolumn{7}{c}{\textbf{\textit{Medical Models}}} \\
\hline
Lingshu-7B~\cite{xu2025lingshu} & 58.89 & 7.13 & 13.94 & 51.94 & 9.52 & 16.31\\
LLaVA-Med-7B~\cite{li2023llava} & 50.02 & 5.21 & 14.95 & 27.71 & 6.56 & 16.45\\
MedVLM-R1-2B~\cite{pan2025medvlm} & 32.26 & 2.10 & 15.51 & 56.58 & 9.05 & 18.27\\
HealthGPT-M3-3.8B~\cite{lin2025healthgpt} & 25.39 & 1.00 & 7.53 & 37.62 & 1.22 & 9.13\\
\rowcolor{lightblue}
\textbf{HeartcareGPT-3.8B} & 61.29 & \textbf{26.84} & \textbf{34.39} & \textbf{78.50} & \underline{23.10} & \underline{38.68}\\
\rowcolor{lightblue}
\textbf{HeartcareGPT-7B} & \textbf{76.55} & \underline{21.70} & \underline{32.55} & 65.03 & \textbf{27.17} & \textbf{44.86}\\
\bottomrule
\end{tabular}
\end{adjustbox}
\vspace{-3mm}
\end{table}

\subsection{Ablation Study and Expert Evaluation}
\label{sec:ablation_study}

We conduct extensive ablation studies to analyze the contribution of each design component in HeartcareGPT. The complete results are presented in Appendix~\ref{app:ablation}.

\begin{figure}[ht]
\centering
\includegraphics[width=0.9\columnwidth]{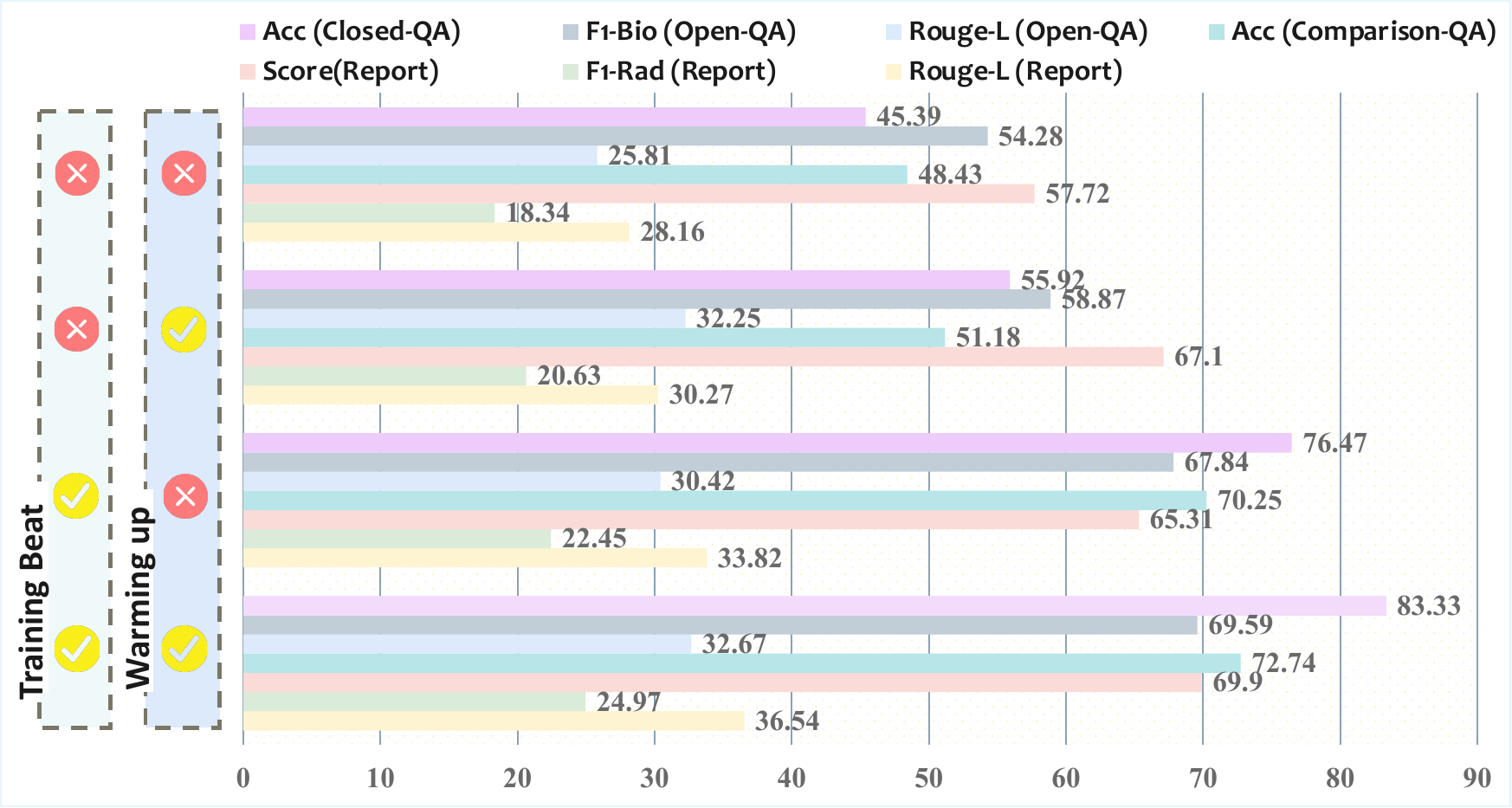}
\caption{Results of ablation studies on training pipeline.}
\label{fig:ablation}
\vspace{-3mm}
\end{figure}

\noindent \textbf{The Three-Stage Training Pipeline.} We further verify the necessity of each training stage in our three-phase optimization scheme. We experiment with models that omit (i) Step 1: beat-level training, (ii) Step 2: projector warming-up, and (iii) both steps simultaneously. Figure~\ref{fig:ablation} show our results of ablation studies on the training pipeline. The performance significantly drops in all ablated variants, indicating that each stage plays an indispensable role in stabilizing multimodal alignment and improving diagnostic accuracy.

\begin{figure}[ht]
\centering
\vspace{-3mm}
\includegraphics[width=0.9\columnwidth]{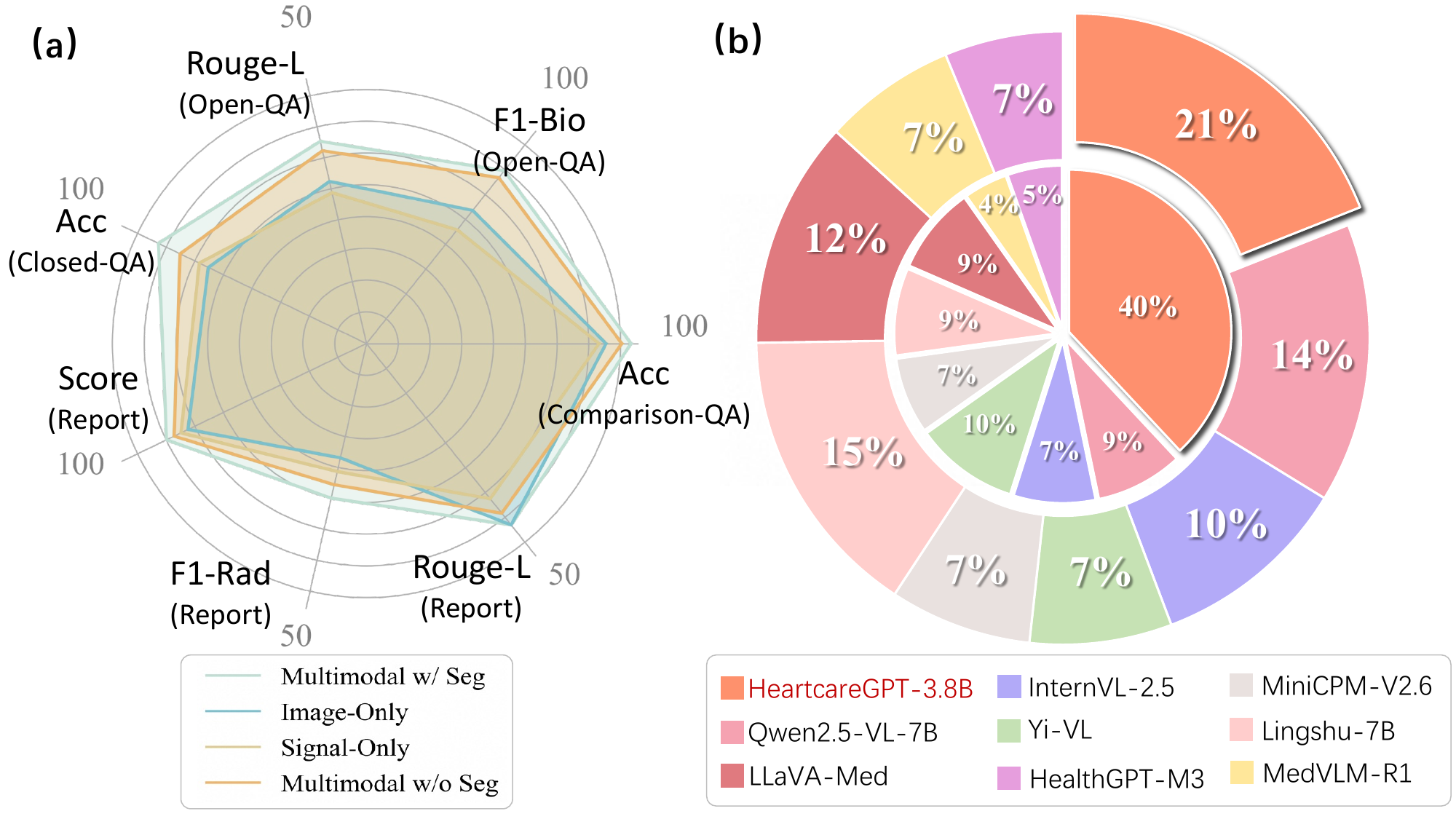}
\caption{(a) Results of ablation studies on multimodal integration. (b) Expert preference distribution across models. Inner-ring results are based on \textit{Open-QA} evaluations; outer-ring results are based on \textit{Report Generation} evaluations.}
\label{fig:expert}
\vspace{-3mm}
\end{figure}

\noindent \textbf{Multimodal Integration.} To assess multimodal fusion, we compare our tri-modal (signal--image--text) model against single-modality baselines (image-only and signal-only). We further examine ECG image segmentation by contrasting full-image and 12-lead sub-image strategies. As summarized in Figure~\ref{fig:expert} (a), multimodal fusion and sub-image segmentation yield clear performance gains, highlighting the value of integrating diverse ECG modalities for improved diagnostic understanding.

\noindent \textbf{Expert Evaluation.} We conduct expert evaluation to assess clinical preference on \textit{Open-QA} and \textit{Report Generation} tasks. Ten board-certified cardiologists reviewed 400 sampled cases with shuffled outputs from HeartcareGPT-3.8B and eight representative baselines, selecting responses best aligned with clinical reasoning and diagnostic conventions. As shown in Figure~\ref{fig:expert} (b), HeartcareGPT achieves the highest first-choice selections---40\% in \textit{Open-QA} and 21\% in \textit{Report Generation}---substantially surpassing all baselines (7--15\%). These results demonstrate clear clinical preference for HeartcareGPT, with expert scoring consistent with automated metrics, further supporting its clinical reliability.
\section{Conclusion}

Heartcare Suite establishes a comprehensive multimodal foundation framework for fine-grained ECG understanding, integrating high-quality dataset, clinically aligned benchmarks, and scalable modeling strategies. We hope this work serves as a stepping stone for future research on Med-MLLMs in clinically grounded signal--language reasoning.
\section*{Acknowledgements}

This work has been supported in part by the NSFC (No. 62436007), the ZJNSF (No. LZ25F020004), the Key Research and Development Projects in Zhejiang Province (No. 2025C01128, 2025C01030, 2025C02156), Ningbo Yongjiang Talent Introduction Programme (2023A400-G).
{
    \small
    \bibliographystyle{ieeenat_fullname}
    \bibliography{main}
}
\newpage
\appendix
\vfill
\cleardoublepage
\begin{center}
\Large \textbf{Appendix}
\end{center}

This is the Appendix for "HeartcareGPT: A Unified Multimodal ECG Suite for Dual Signal--Image
Modeling and Understanding".

This Appendix is organized as follows:

\begin{itemize}
\item Section~\ref{app:engine} provides the details of our proposed multimodal data engine---\textbf{HeartAgent}.
\item Section~\ref{app:implement} provides the details of the experimental implementation, the training process of \textbf{HeartcareGPT}, the construction details of \textbf{Heartcare-400K}, and the specific information of \textbf{Heartcare-Bench}.
\item Section~\ref{app:results} shows our detailed ablation experimental results of \textbf{HeartcareGPT} and \textbf{Beat}.
\item Section~\ref{app:case} shows typical data examples in \textbf{Heartcare-400K}.
\item Section~\ref{app:limit} lists the broader impact and limitations of this paper.
\end{itemize}

\section{Multimodal Data Engine Details}
\label{app:engine}

To efficiently construct the Heartcare-400K dataset, we design \textbf{HeartAgent}, an automated multimodal data engine that transforms multi-source ECG data into high-quality, structured QA pairs. As illustrated in Figure~\ref{fig:engine}, the system consists of four core modules that work collaboratively to complete the full pipeline from raw data parsing to task QA generation.

\noindent \textbf{Multimodal Feature Converter.} The Converter preprocesses hospital PDF reports into standardized inputs for downstream modules. It extracts patient metadata (e.g., age and gender) and diagnostic details (diagnosis, waveform, rhythm and other features) from the PDF via fitz (PyMuPDF)~\cite{pymupdf} and regular expressions. Subsequently, diagnostic text is mapped to structured English labels in accordance with the SCP-ECG semantic standard~\cite{rubel2016scp}.

\noindent \textbf{Noise Filtering and Quality Optimizer.} The Optimizer implements a parsing mechanism to unify heterogeneous ECG inputs into standardized 12-lead, 250 Hz digital signals. Structured digital inputs from public datasets are processed via the WFDB toolkit~\cite{wfdbapptoolbox}. To correct noise, drift, and missing-segment issues, the Optimizer applies a three-stage enhancement pipeline: (i) resampling all signals to 250 Hz, (ii) applying clean functions in NueroKit2~\cite{neurokit2} for lead-level denoising and baseline correction, and (iii) using NeuroKit2’s quality scores to extract high-quality segments while discarding low-quality regions.

\noindent \textbf{Diversed Image Generator.} The Image Generator processes clear waveform images through a dual-channel pipeline. For high-quality signals originated from the Optimizer, the Generator uses Matplotlib~\cite{hunter2007matplotlib} in the final stage to render clean digital signals into images. For hospital reports, it processes PDF-based reports to produce cropped images of the complete 12-lead tracings, while ensuring all data is de-identified.

\noindent \textbf{Multi-Task QA Builder.} To enhance model generalization and training consistency across multi-level ECG QA scenarios, we use GPT-4~\cite{gpt4v} to directly generate structured multimodal ECG QA samples, constructing the training dataset. Each generated sample consists of four key components: \textbf{(i) Context Description}, providing background information such as patient demographics, signal snippets, or clinical report summaries; \textbf{(ii) Task Instruction}, specifying the required operation type, including \textit{Closed-QA}, \textit{Open-QA}, \textit{Comparison-QA}, \textit{Report Generation} and \textit{Signal Prediction}; \textbf{(iii) Auxiliary Labels}, providing additional structured supervisory information to enhance training quality; and \textbf{(iv) Output Format}, standardizing expression of answers to ensure consistency across tasks. 

\section{Implementation Details}
\label{app:implement}

\subsection{Model Details}
\label{app:model}

HeartcareGPT employs an architecture design that aligns ECG signals and images with textual inputs in latent space. For signal and image modalities, We use two MLP projectors for cross-modal feature fusion. Notably, we implement LoRA~\cite{hu2022lora} for parameter-efficient fine-tuning, preserving pretrained knowledge while enabling domain-specific adaptation for ECG tasks. This design achieves an optimal balance between model capacity and computational efficiency, establishing a scalable architectural foundation for multimodal ECG understanding.

\begin{table*}[ht]
\centering
\renewcommand{\arraystretch}{1.2}
\caption{Overview of the components of HeartcareGPT.}
\label{tab:components}
\begin{adjustbox}{width=0.98\textwidth,center}
\begin{tabular}{lccccccc}
\toprule
\textbf{Model} & \textbf{ViT} & \textbf{Signal Projector} & \textbf{Image Projector} & \textbf{LLM} & \textbf{Params} & \textbf{Vocab Size} & \textbf{LoRA Rank}\\
\midrule
\textbf{HeartcareGPT-3.8B} & SigLIP-So400M-patch14-384 & 1-layer MLP & 2-layer MLP & Phi-3-mini-4k-Instruct & 3.8B & 32,069 & 64   \\
\textbf{HeartcareGPT-7B} & SigLIP-So400M-patch14-384 & 1-layer MLP & 2-layer MLP & Qwen2.5-7B-Instruct & 7B & 151,851 & 64   \\
\bottomrule
\end{tabular}
\end{adjustbox}
\end{table*}

HeartcareGPT offers two versions: \textbf{HeartcareGPT-3.8B} and \textbf{HeartcareGPT-7B}, which are based on Phi-3-mini-4k-Instruct~\cite{abdin2024phi} and Qwen2.5-7B-Instruct~\cite{bai2025qwen2} as pre-trained LLMs, respectively. We employs SigLip-So400M-patch14-384~\cite{zhai2023sigmoidlosslanguageimage} and the proposed Beat as dual-form feature encoders, and extend the model’s native textual tokenizer by adding four special tokens: $\langle \texttt{PRED}\rangle$, $\langle \texttt{sig}\rangle$, $\langle \texttt{img}\rangle$ and $\langle \texttt{text}\rangle$. $\langle \texttt{PRED}\rangle$ indicates that the model is expected to perform signal-prediction tasks handled by the integrated beat module. $\langle \texttt{sig}\rangle$ marks the starting boundaries of sequential features, $\langle \texttt{img}\rangle$ marks the starting boundaries of visual features, and $\langle \texttt{text}\rangle$ marks the starting boundaries of textual features, enabling the model to more reliably recognize and process these modality-specific representations. Table~\ref{tab:components} shows the details.

In addition, for baseline models that support only a single input modality, we apply a unified preprocessing strategy to align modalities, ensuring fairness within the same evaluation framework. These baseline models typically support only image inputs rather than raw signals. Therefore, we convert the ECG signals into images using the Diverse Image Generator module of HeartAgent (presented in Appendix~\ref{app:engine}), and feed the generated images together with the textual questions into the baseline models to obtain their responses.

\subsection{Training Details}
\label{app:train}

We follow a three-stage training paradigm: (i) training Beat to extract high-fidelity ECG embeddings, (ii) warming up the visual and signal projectors to stabilize feature alignment, and (iii) performing joint instruction fine-tuning on Heartcare-400K for end-to-end modeling. This paradigm achieves decoupled feature learning and semantic alignment across stages, enabling the model to maintain signal fidelity while acquiring advanced clinical reasoning capabilities.

\noindent \textbf{Training Beat.} Beat is first trained on PTB-XL~\cite{wagner2020ptb} dataset. We use a joint supervision strategy to optimize reconstruction and prediction losses simultaneously. This stage focuses on learning robust ECG signal representations through DVQ structure.

\noindent \textbf{Warming Up.} We warm up the visual and signal projectors using paired ECG images and signals, allowing the model to align heterogeneous features in a controlled manner. By isolating projector optimization, the model avoids early-stage instability and catastrophic feature distortion. This step establishes a coherent multimodal latent space for unified processing.

\noindent \textbf{Joint Fine-Tuning.} We perform full-model instruction fine-tuning on Heartcare-400K, enabling the model to associate ECG inputs with diagnostic reasoning, structured reporting, and question answering. The model learns both high-level clinical knowledge and fine-grained ECG interpretation skills. This stage integrates all modalities end-to-end, yielding a clinically aligned and diagnostically capable ECG foundation model.

\begin{table*}[ht]
\centering
\renewcommand{\arraystretch}{1.2}
\caption{Overview of hyperparameter configurations.}
\label{tab:parameter}
\begin{adjustbox}{width=0.98\textwidth,center}
\begin{tabular}{lcccccc}
\toprule
\multirow{2}{*}{\textbf{Hyperparameter}} & \multicolumn{3}{c}{\textbf{HeartcareGPT-3.8B}} & \multicolumn{3}{c}{\textbf{HeartcareGPT-7B}} \\
\cmidrule(lr){2-4} \cmidrule(lr){5-7}
& \textbf{Training Beat} & \textbf{Warming up} & \textbf{Joint Fine-Tuning} & \textbf{Training Beat} & \textbf{Warming up} & \textbf{Joint Fine-Tuning}\\
\midrule
\textbf{Optimizer} & AdamW & AdamW & AdamW & AdamW & AdamW & AdamW \\
\textbf{Adapter LR} & / & / & 2e-5 & / & / & 2e-5\\
\textbf{Learning Rate} & 1e-4 & 2e-4 & 2e-5 & 1e-4 & 2e-4 & 2e-5\\
\textbf{Global Batch Size} & 32 & 16 & 16 & 32 & 16 & 16 \\
\textbf{Weight Decay} & 0 & 0.01 & 0.01 & 0 & 0.01 & 0.01\\
\textbf{Dropout Rate} & 0 & 0 & 0.05 & 0 & 0 & 0.05\\
\textbf{LR Scheduler} & Cosine Annealing & Linear & Linear & Cosine Annealing & Linear & Linear\\
\bottomrule
\end{tabular}
\end{adjustbox}
\end{table*}

Hyperparameter configurations for each training stage are detailed in Table~\ref{tab:parameter}.

\subsection{Construction details of Heartcare-400K}
\label{app:dataset}

\noindent \textbf{Data Source Details.} In the data collection phase, we gather ECG report data with two modalities---digitized raw signals and clinical report images.

PTB-XL~\cite{wagner2020ptb} is one of the largest publicly available electrocardiogram datasets, comprising 21,799 clinical 12-lead ECG recordings that cover a diverse range of cardiac pathologies as well as healthy control data. Each recording has a duration of 10 seconds with a sampling rate of 500 Hz, accompanied by standardized diagnostic annotations and detailed patient metadata, such as gender and age. We utilize PTB-XL as a high-quality structured data source to enhance the diversity and accuracy of Heartcare-400K in the digital modality.

In contrast, ECG image modality data has long been constrained by acquisition challenges, annotation costs, and privacy concerns, resulting in scarce and outdated publicly available image datasets. To address this issue, we establish collaborations with two top-tier hospitals and collect a total of 12,170 recent ECG report forms through rigorous anonymization and professional physician annotations. Each report is in a standardized PDF format, containing basic patient information, physiological parameters, physician diagnoses, and approximately 5-second 12-lead image recordings, significantly improving the timeliness and clinical usability of the image modality.

\begin{figure}[ht]
\centering
\includegraphics[width=0.98\columnwidth]{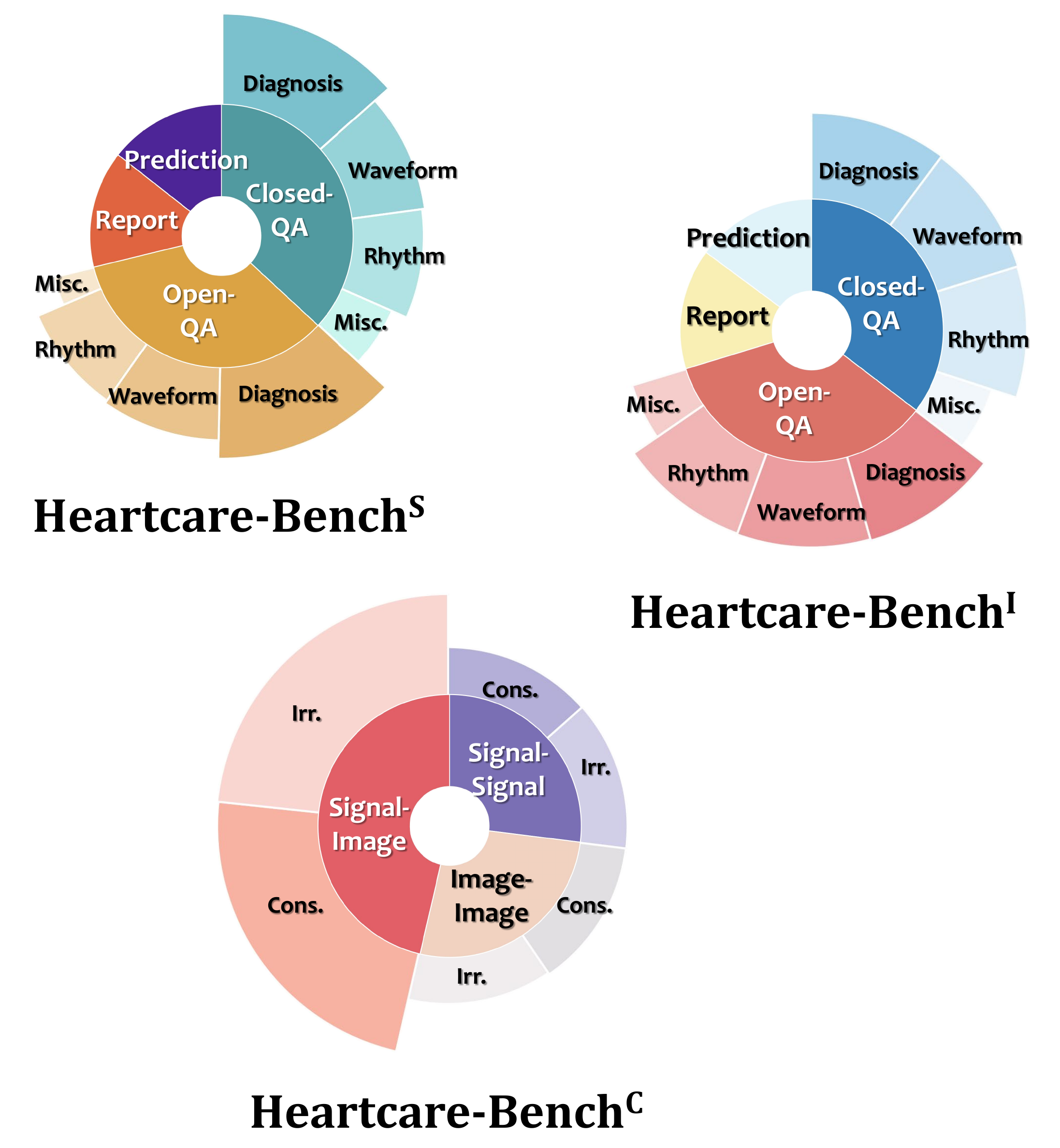}
\caption{Hierarchical distribution of Heartcare-400K dataset across benchmarks and subtask types.}
\label{fig:dataset}
\end{figure}

To provide a comprehensive overview of diagnostic coverage and clinical relevance of Heartcare-400K, Table~\ref{tab:category} presents a detailed listing of ECG features organized across four clinically critical dimensions: (i) \textit{Diagnosis} (e.g., Inferior MI, AV block), (ii) \textit{Waveform abnormalities} (e.g., T abnormalities, ST depression), (iii) \textit{Rhythm patterns} (e.g., sinus rhythm, atrial fibrillation), and (iv) \textit{Miscellaneous features} (e.g., noise, heart axis). This illustrates the dataset’s comprehensive and balanced representation across these clinically relevant dimensions.

To visually summarize the dataset composition and task-specific coverage of Heartcare-400K, we provide a hierarchical radial visualization in Figure~\ref{fig:dataset}. This depict the distribution of QA subtasks within three benchmarks, Heartcare-Bench\textsuperscript{S}, Heartcare-Bench\textsuperscript{I} and Heartcare-Bench\textsuperscript{C}, highlighting both the relative scale and fine-grained breakdown of different data types.

\noindent \textbf{QA Templates.} For datasets that only contain classification or grading labels, we analyze the data characteristics of their labels and design different QA templates for each. This allow us to transform the original data into QA pairs. We additionally provide the prompt used to guide GPT-4~\cite{gpt4v} in generating QA templates for diagnosis subtasks in \textit{Closed-QA} data construction pipeline of HeartAgent, as shown in Figure~\ref{fig:qa_prompt}. Examples of the QA templates are shown in the Table~\ref{tab:qa_templates}.

\subsection{Construction details of Heartcare-Bench}
\label{app:bench}

\arrayrulecolor{black}

\begin{table}[ht]
\centering
\renewcommand{\arraystretch}{1.1}
\caption{Evaluation dimensions and weighted criteria for ECG diagnostic reports.}
\label{tab:eval_criteria}
\begin{adjustbox}{width=0.98\columnwidth,center}
\begin{tabular}{p{1.8cm}p{6.7cm}c}
\toprule
\textbf{Dimension} & \textbf{Evaluation Criteria} & \textbf{Weight} \\
\midrule

\multirow{6}{1.2cm}{Diagnosis Completeness} 
& Completeness of abnormal features mentioned & 10 \\
\cmidrule(l){2-3}
& Completeness of key diagnoses included & 10 \\
\cmidrule(l){2-3}
& Absence of critical diagnostic errors & 10 \\
\cmidrule(l){2-3}
& Whether the report describes severity or likelihood of the findings & 8 \\
\cmidrule(l){2-3}
& Whether the report includes suspected diagnoses & 7 \\
\midrule

\multirow{4}{1.2cm}{Waveform Accuracy}
& Correct identification of anatomical regions (e.g., P/QRS/T waves) & 8 \\
\cmidrule(l){2-3}
& Correct recognition of waveform abnormalities (e.g., ST elevation/depression) & 7 \\
\midrule

\multirow{7}{1.2cm}{Rhythm Accuracy}
& Correct classification of baseline rhythm (e.g., sinus or ectopic) & 4 \\
\cmidrule(l){2-3}
& Correct classification of arrhythmias (e.g., tachycardia or bradycardia) & 4 \\
\cmidrule(l){2-3}
& Correct interpretation of conduction abnormalities (e.g., location and degree of block) & 4 \\
\cmidrule(l){2-3}
& Accurate detection of pacing signals & 3 \\
\midrule

\multirow{6}{1.2cm}{Report Logic}
& Report is well-structured and logically organized & 5 \\
\cmidrule(l){2-3}
& Findings are explained in a point-wise or categorized manner & 4 \\
\cmidrule(l){2-3}
& Includes relevant auxiliary information (e.g., age, gender, etc.) & 3 \\
\cmidrule(l){2-3}
& Patient privacy is protected via anonymization & 3 \\
\midrule

\multirow{5}{1.2cm}{Descriptive Norms}
& Terminology complies with SCP-ECG standards (e.g., use "complete right bundle branch block" instead of "RBBB") & 5 \\
\cmidrule(l){2-3}
& Language avoids inappropriate certainty (e.g., avoids overconfident conclusions) & 5 \\
\midrule

\textbf{Total Score} & & \textbf{100} \\
\bottomrule
\end{tabular}
\end{adjustbox}
\end{table}

To comprehensively assess model performance across different task types, we design a multi-dimensional evaluation framework tailored to the specific objectives of each module. The evaluation criteria are carefully selected to reflect the core competencies required by each task---ranging from answer correctness and semantic understanding to clinical accuracy and waveform forecasting fidelity.

\noindent \textbf{Closed‐QA.} We measure model discrimination performance by standard \textit{Accuracy}.

\noindent \textbf{Open‐QA.} We adopt a dual‐track evaluation comprising \textit{F1-Bio}~\cite{ramshaw1995text} to assess semantic alignment, and \textit{ROUGE-L}~\cite{lin2004rouge} to quantify linguistic fluency and contextual style fidelity.

\noindent \textbf{Comparision‐QA.} We measure model discrimination performance by standard \textit{Accuracy}.

\noindent \textbf{Report Generation.} (i) Beyond ROUGE-L, we use \textit{F1-RadGraph}~\cite{jain2021radgraph} to evaluate the precision of entity and relation extraction in the report structure. (ii) We implement an 100-point, multi-dimensional criteria that evaluates diagnostic completeness, waveform accuracy, rhythm accuracy, report logic, and descriptive norms. Using GPT-4~\cite{gpt4v}, we tally error types and severity according to this criteria (see Table~\ref{tab:eval_criteria}) for weighted penalties. The template used for GPT-4’s evaluation is shown in Figure~\ref{fig:eval_prompt}. According to the evaluation criteria, we grade the reports as follows:

\begin{itemize}
\item \textbf{Excellent Report (90--100):} Nearly no error with complete diagnostic information, clear structure, and no clinically significant mistakes. Ready for immediate clinical use.
\item \textbf{Acceptable Report (80--89):} Contains minor errors but maintains diagnostic accuracy and logical flow. Requires minimal editing before clinical application.
\item \textbf{Review Required Report (60--79):} Has notable errors, incomplete information, or unclear structure. Needs expert verification before use.
\item \textbf{Unusable Report (< 60):} Contains critical errors, major missing information, or serious diagnostic inaccuracies. Unsafe for clinical decision-making.
\end{itemize}

\noindent \textbf{Signal Prediction.} We delimit the predicted segment the with special token $\langle \texttt{PRED}\rangle$ and compute the \textit{Mean Squared Error (MSE)} between the forecasted waveform and the true continuation. Lower MSE indicates superior prediction accuracy.

Heartcare-Bench adopts strict patient-level partitioning to ensure that all ECG records from the same individual---across different formats or tasks---reside in a single split. To further eliminate potential leakage, we perform a lightweight yet comprehensive overlap check that includes examination-level matching using study IDs and timestamps, duplicate detection of ECG waveforms via normalized hashing, and visual and textual duplication inspection for rendered ECG images and associated reports. These checks jointly guarantee that no record or derived annotation appears across splits, ensuring fair and leakage-free evaluation.

\section{Supplemental Experimental Results}
\label{app:results}

\subsection{Ablation Study Results of HeartcareGPT}
\label{app:ablation}

For completeness, we present in this appendix the full numerical results of all ablation settings of HeartcareGPT-3.8B discussed in Section~\ref{sec:ablation_study}. While the main paper visualizes the trends through figures, the detailed data tables are provided here to enable transparent comparison across all variants of the training pipeline (shown in Table~\ref{tab:stage_closed},~\ref{tab:stage_open},~\ref{tab:stage_comparsion} and~\ref{tab:stage_report}) and multimodal integration strategies (shown in Table~\ref{tab:multi_closed},~\ref{tab:multi_open},~\ref{tab:multi_comparsion} and~\ref{tab:multi_report}).

Across all settings, the ablation results consistently validate the effectiveness of our design choices. Removing either Beat training stage or warming up stage leads to substantial degradation in diagnostic accuracy and multimodal alignment. Likewise, replacing our tri-modal fusion with single-modality variants or removing 12-lead sub-image segmentation yields clear declines in performance. These trends reinforce the necessity of ECG-aware representation modeling, stabilized multimodal projection, and fine-grained visual decomposition for enhancing clinical reasoning.

Overall, experimental results further substantiate our claims that each architectural and training design contributes meaningfully to HeartcareGPT’s superior diagnostic capability and its robustness across diverse ECG modalities.

\subsection{Ablation Study Results of Beat}
\label{app:beat}

To validate the performance of our model, we conduct comprehensive experiments based on our ECG tokenizer, Beat. We evaluate its capabilities in both signal reconstruction and prediction tasks under various structural configurations, including the use of DVQ structure, codebook size, and input sequence length.

In Section~\ref{sec:beat}, we have introduced the joint supervision strategy adopted in Beat. To facilitate evaluation in experiments, we also consider this total loss as a quantitative metric. Specifically, the overall training objective of Beat $\mathcal{L}_{\text{total}}$ is defined as:

\begin{equation}
\mathcal{L}_{\text{total}} = \lambda_1 \mathcal{L}_{\text{recon}} + \lambda_2 \mathcal{L}_{\text{pred}} + \lambda_3 \mathcal{L}_{\text{VQ}},    
\end{equation}

where we set $\lambda_1=1.0$, $\lambda_2=0.5$ and $\lambda_3=0.25$ in our experiments, balancing the contributions of reconstruction, prediction, and vector-quantization objectives. By monitoring the overall training objective as well as its individual components, we can more thoroughly assess how each architectural modification affects representation quality and optimization stability.

For a more comprehensive evaluation of Beat's performance, we additionally employ the codebook utilization as $\text{Code}\%$, which quantifies the proportion of codebook entries actually used during encoding.

\begin{table}[ht]
\centering
\renewcommand{\arraystretch}{1.2}
\caption{Ablation study for Beat under different configurations.}
\label{tab:tokenizer}
\begin{adjustbox}{width=0.9\columnwidth,center}
\begin{tabular}{ccc|cc}
\toprule
\textbf{DVQ} & \textbf{Codebook Size} & \textbf{Input Length} & \textbf{$\text{Code}\%$} & \textbf{$\mathcal{L}_{\text{total}}$}\\
\midrule
- & 512 & 1000 & 93.75 & 1.00\\
\faCheckCircle & 1024 & 1000 & 79.50 & \underline{0.79}\\
\faCheckCircle & 256 & 1000 & \textbf{99.87} & 0.81\\
\faCheckCircle & 512 & 1500 & 91.46 & 1.10\\
\faCheckCircle & 512 & 500 & \underline{98.20} & 0.83\\
\rowcolor{lightblue}
\faCheckCircle & 512 & 1000 & 96.22 & \textbf{0.76}\\
\bottomrule
\end{tabular}
\end{adjustbox}
\end{table}

As shown in Table~\ref{tab:tokenizer}, the results show that the DVQ structure with a codebook size of 512 and an input length of 1000 strikes a favorable balance between compression efficiency and semantic completeness. Notably, while smaller codebooks or shorter input sequences can achieve higher codebook utilization, their overall loss is higher than our chosen configuration, highlighting that code utilization alone is not sufficient to evaluate tokenizer effectiveness.

We summarize the following observations: (i) The DVQ structure captures global rhythm patterns via the core codebook and refines local variations via the residual codebook, thereby enhancing the clinical semantic integrity of the discrete representation while maintaining a compact token space. (ii) Enlarging the codebook increases representational granularity but leads to codebook collapse and lower utilization, whereas a smaller codebook fails to capture the complex pathological semantics of ECG signals. (iii) Excessively long or short input sequences degrade codebook utilization and introduce instability in reconstruction and prediction, likely due to imbalanced temporal context or fragmented signal structure. Overall, Beat achieves an effective global-local modeling trade-off through structural and parametric design, significantly improving the quality of ECG tokenization and enabling end-to-end training of ECG and text modalities within Med-MLLMs.

\begin{figure*}[ht]
\centering
\includegraphics[width=0.9\textwidth]{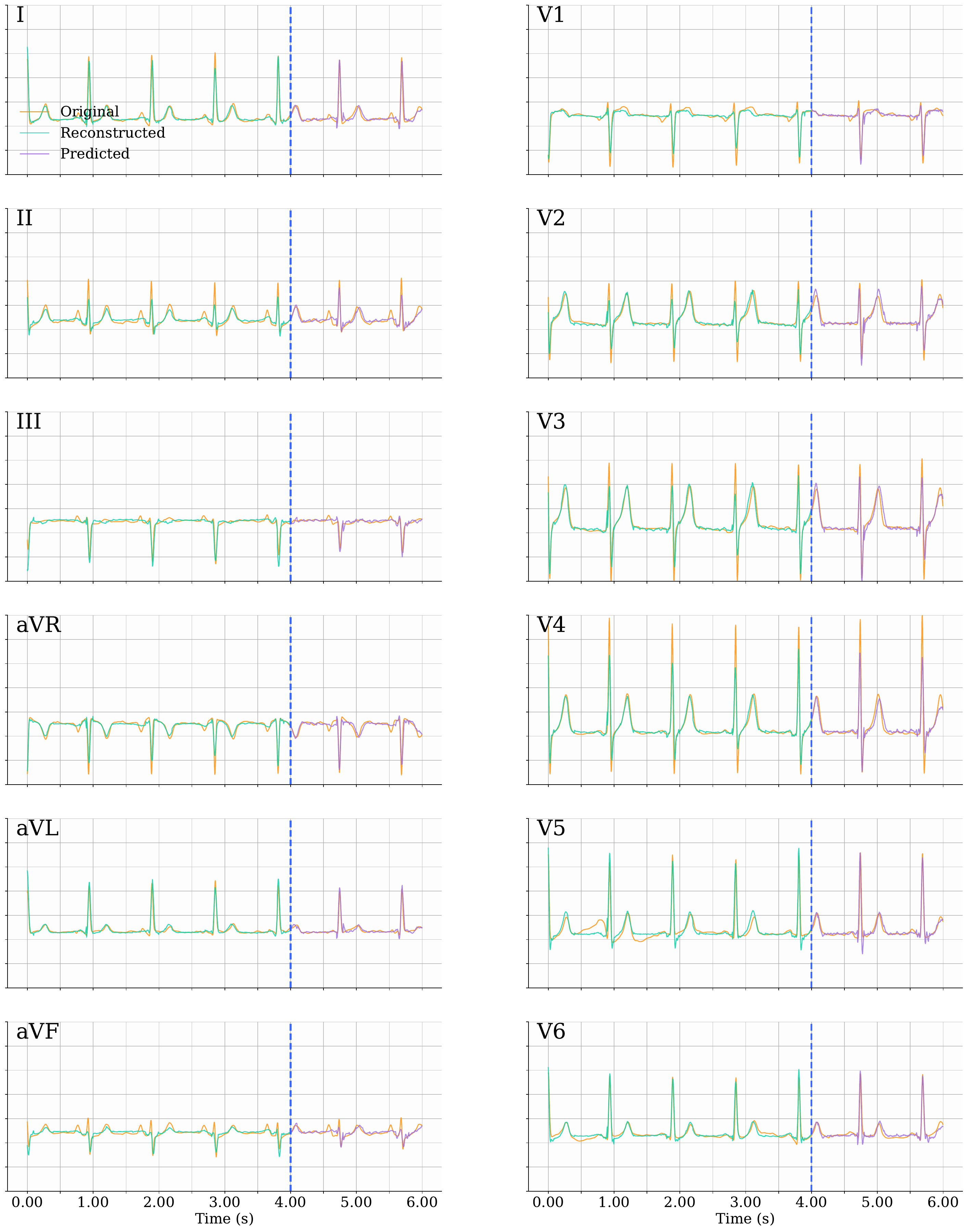}
\caption{ECG signal reconstruction and prediction with Beat.}
\label{fig:ECG_signal}
\end{figure*}

Furthermore, Figure~\ref{fig:ECG_signal} provides a comprehensive visualization of Beat's reconstruction and prediction performance, demonstrating the model's capability to accurately recover input patterns while generating high-fidelity future predictions.

\section{Case Study}
\label{app:case}

\begin{figure*}[ht]
\centering
\includegraphics[width=0.98\textwidth]{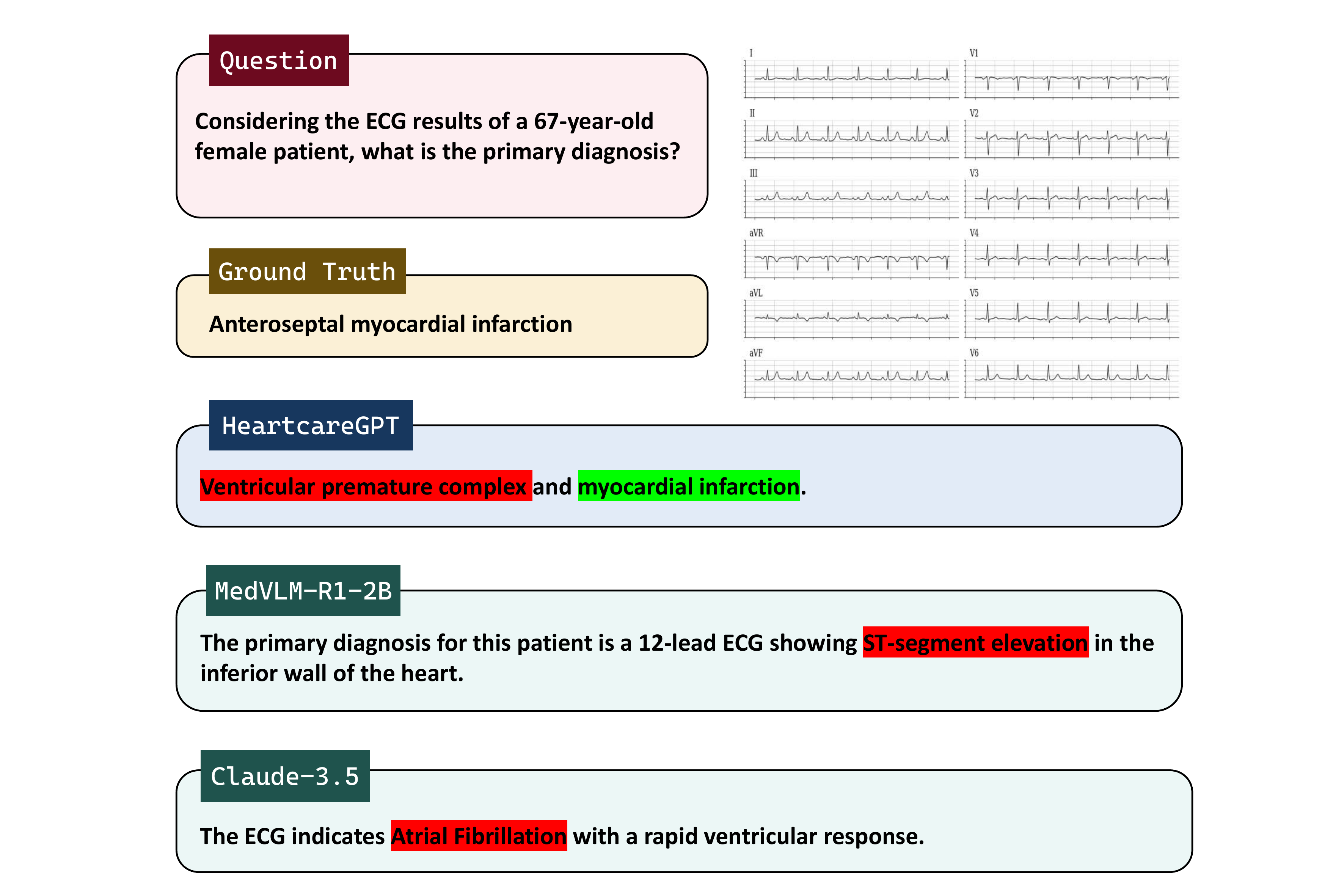}
\caption{A case of \textit{open-QA} tasks.}
\label{fig:case_open}
\end{figure*}

\begin{figure*}[ht]
\centering
\includegraphics[width=0.98\textwidth]{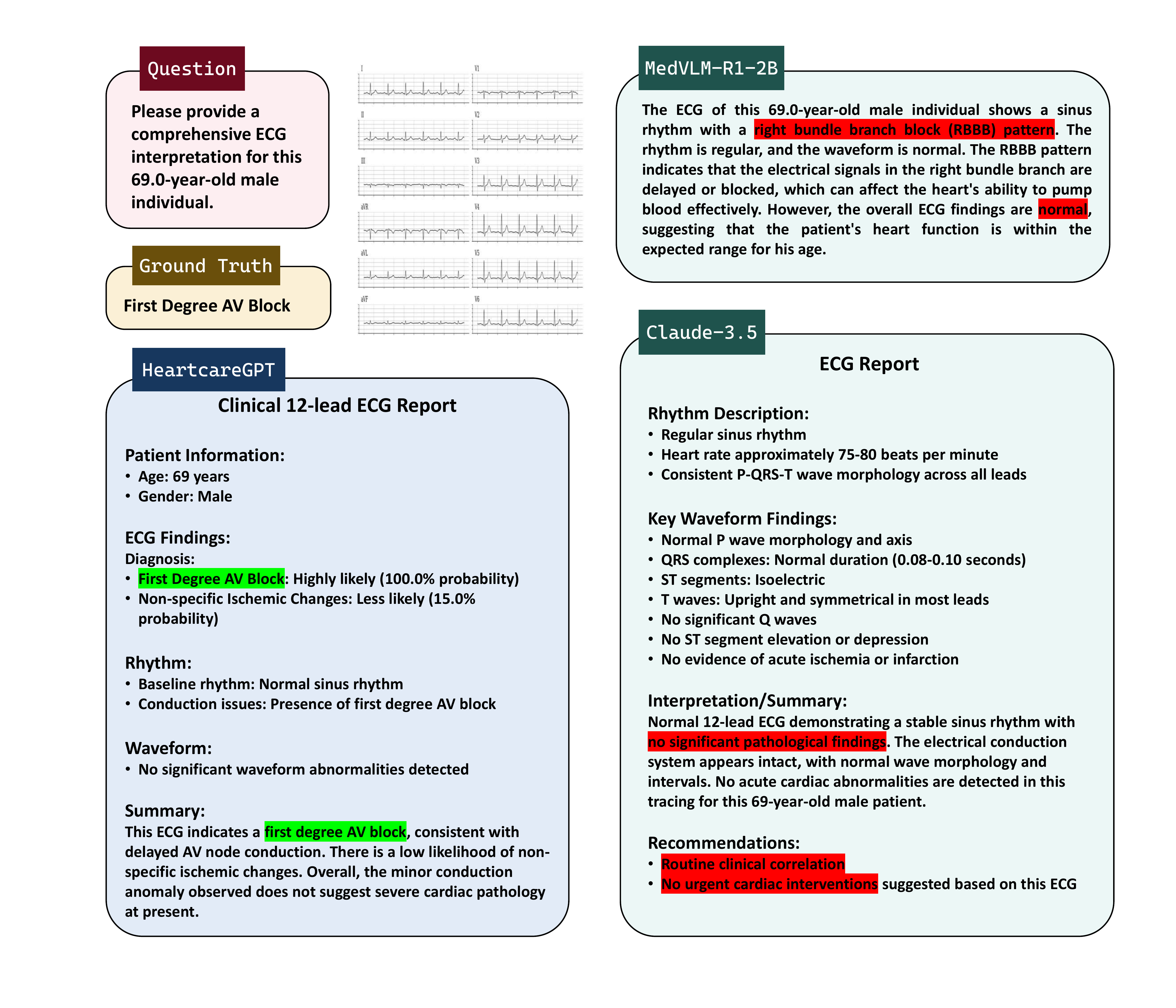}
\caption{A case of \textit{Report Generation} tasks.}
\label{fig:case_report}
\end{figure*}

In this section, we compare generated answers of our proposed HeartcareGPT-3.8B with those of a generalist model (Claude-3.5) and a medical model (MedVLM-R1-2B). Since \textit{Closed-QA} and \textit{Comparison-QA} primarily contain selection-based questions, which are less illustrative for qualitative comparison, we focus on \textit{Open-QA} and \textit{Report Generation} cases where the model’s clinical reasoning capabilities can be more clearly observed. Figures~\ref{fig:case_open} and~\ref{fig:case_report} illustrate the performance of these three models on \textit{Open-QA} and \textit{Report Generation} tasks. We highlight statements that match the ground truth in \colorbox{green}{green}, and indicate discrepancies in \colorbox{red}{red}.

As shown in Figure~\ref{fig:case_open}, HeartcareGPT’s response closely matches the reference answer, showing its precise understanding of detailed diagnostic questions.

A similar observation can be made from Figure~\ref{fig:case_report}. HeartcareGPT produces reports that not only capture clinically relevant findings with high accuracy, but also maintain clear structure. Moreover, the reports show correct usage of ECG-specific phrasing and logical flow.

\section{Limitations}
\label{app:limit}

Heartcare Suite advances multimodal ECG understanding with potential benefits for clinical diagnosis, medical artificial intelligence (AI) research, and patient care. By integrating raw ECG signals and structured reports, it enables accurate, automated cardiac analysis, particularly valuable in resource-limited settings. The release of Heartcare-400K and Heartcare-Bench fosters transparency and progress in medical AI. However, limitations include dataset biases (e.g., underrepresentation of rare conditions), potential signal fidelity loss in tokenization, and untested real-time monitoring capabilities. Computational costs and regulatory hurdles for clinical deployment remain challenges. Future work should expand data diversity, optimize real-time processing, and validate clinical utility through trials.

\begin{table*}[ht]
\centering
\renewcommand{\arraystretch}{1.1}
\caption{Sample QA templates in Heartcare-400K.}
\label{tab:qa_templates}
\begin{adjustbox}{width=0.95\textwidth,center}
\begin{tabular}{p{2.5cm}p{17cm}}
\hline

\rowcolor{gray!10}
\multicolumn{2}{c}{\textbf{Questions}} \\
\hline

\multirow{14}{2.5cm}{\textbf{Closed-QA}}
& 1. Please assign the most suitable shape and structure classification with a detailed examination of the provided ECG sequence of this subject. \\ 
& A. Non-diagnostic T abnormalities; B. Ventricular premature complex; C. Low QRS voltage in limb leads; D. Non-specific ST elevation. \\
\cmidrule(lr){2-2}

& 2. Investigate the patient's ECG reading and diagnose its classification based on its features. \\
& A. Normal ECG; B. Incomplete left bundle branch block; C. Long QTc-interval; D. Complete right bundle branch block. \\
\cmidrule(lr){2-2}

& 3. By conducting a detailed evaluation of the ECG trace of the person, output the correct rate and regularity it should be classified under. \\
& A. Bigeminal pattern; B. Sinus tachycardia; C. Sinus rhythm; D. Normal functioning artificial pacemaker. \\
\cmidrule(lr){2-2}

& 4. What would you determine the pattern and timing of this ECG reading to be? \\
& A. Atrial fibrillation; B. Atrial flutter; C. Normal functioning artificial pacemaker; D. No abnormality detected. \\
\cmidrule(lr){2-2}

& 5. With precision and attention to detail, work through the subject's ECG reading and give the most appropriate rhythm based on its characteristics. \\
& A. Sinus bradycardia; B. Atrial flutter; C. Paroxysmal supraventricular tachycardia; D. Atrial fibrillation. \\
\midrule

\multirow{6}{2.5cm}{\textbf{Open-QA}}
& 1. Given the ECG finding, please work through its features and classify the right shape and structure. \\ 
\cmidrule(lr){2-2}
& 2. Assign the waveform associated with the ECG characteristic. \\ 
\cmidrule(lr){2-2}
& 3. What pattern and timing does ECG interpretation exhibit? \\ 
\cmidrule(lr){2-2}
& 4. Through meticulous examination of the patient's ECG sequence, please accurately determine the diagnosis that best defines it. \\ 
\cmidrule(lr){2-2}
& 5. What rhythm does the given ECG characteristic from the patient exhibit? \\ 
\midrule

\multirow{5}{2.5cm}{\textbf{Comparison-QA}}
& 1. Has non-diagnostic T abnormalities been eliminated in the recent tracing in comparison to the previous one? \\ 
\cmidrule(lr){2-2}
& 2. Does the second ECG still indicate the presence of ST/T change as compared to the first ECG? \\ 
\cmidrule(lr){2-2}
& 3. Is atrial flutter still not detected in the recent tracing when compared to the previous one? \\ 
\cmidrule(lr){2-2}
& 4. Does the second ECG still show the absence of non-specific T-wave changes when compared to the first ECG? \\ 
\cmidrule(lr){2-2}
& 5. Is the RR interval still considered normal in the latest tracing when compared to the previous tracing? \\ 
\midrule

\multirow{5}{2.5cm}{\textbf{Report Generation}}
& 1. Please provide a comprehensive ECG interpretation for this 69.0-year-old male individual. \\ 
\cmidrule(lr){2-2}
& 2. Produce a detailed ECG analysis for this 55.0-year-old female case. \\ 
\cmidrule(lr){2-2}
& 3. Generate a standardized ECG assessment report for this 36.0-year-old male patient. \\ 
\cmidrule(lr){2-2}
& 4. Based on the ECG signal from this 75.0-year-old male patient, generate a structured clinical 12-lead ECG report. \\ 
\cmidrule(lr){2-2}
& 5. Analyze the ECG tracing and generate a clinical report for this 77.0-year-old male individual. \\ 
\hline

\rowcolor{gray!10}
\multicolumn{2}{c}{\textbf{Answers}} \\
\hline

\multirow{5}{2.5cm}{\textbf{Positive Condition}}
& 1. Based on the ECG pattern, after thorough examination, the form is classified as \{condition\}. \\ 
\cmidrule(lr){2-2}
& 2. The diagnostic classification observed in the given ECG observation suggests a evident link to suggestive of \{condition\}. \\ 
\cmidrule(lr){2-2}
& 3. After systematic analysis, the ECG evaluation is classified as \{condition\}. \\ 
\cmidrule(lr){2-2}
& 4. Clinical findings from this ECG assessment reinforce the presence of \{condition\} as a evident outcome. \\ 
\cmidrule(lr){2-2}
& 5. The ECG signal shows evidence of \{condition\}. \\ 
\midrule

\multirow{5}{2.5cm}{\textbf{Negative Condition}}
& 1. All leads demonstrate physiological waveforms, and the overall conclusion is a normal ECG. \\ 
\cmidrule(lr){2-2}
& 2. Standard diagnostic criteria confirm that the signal is entirely normal, with no pathological findings. \\ 
\cmidrule(lr){2-2}
& 3. No evidence of ST-segment elevation, depression, or T-wave inversions. \\ 
\cmidrule(lr){2-2}
& 4. Healthy cardiac activity. \\ 
\cmidrule(lr){2-2}
& 5. Heart rate is regular, with consistent P-P and R-R intervals. \\ 
\bottomrule
\end{tabular}
\end{adjustbox}
\end{table*}
\begin{table*}[ht]
\centering
\renewcommand{\arraystretch}{1.2}
\caption{Systematic Categorization of ECG Features in Heartcare-400K.}
\label{tab:category}
\begin{adjustbox}{width=0.8\textwidth,center}
\begin{tabular}{m{8cm}m{8cm}}
\hline
\rowcolor{gray!10}
\multicolumn{2}{c}{\textbf{Diagnosis}} \\
\hline
ischemic in inferior leads & non-specific ischemic \\
septal hypertrophy & non-specific intraventricular conduction block \\
subendocardial injury in anterolateral leads & left anterior fascicular block \\
ischemic in anterior leads & non-diagnostic T abnormalities \\
anterolateral myocardial infarction & incomplete right bundle branch block \\
non-specific ST changes & third degree AV block \\
right ventricular hypertrophy & ischemic in lateral leads \\
incomplete left bundle branch block & long QTc-interval \\
first degree AV block & inferoposterior myocardial infarction \\
right atrial hypertrophy & Wolf-Parkinson-White syndrome \\
subendocardial injury in anteroseptal leads & inferolateral myocardial infarction \\
inferior myocardial infarction & posterior myocardial infarction \\
right atrial overload/enlargement & long QT-interval \\
complete left bundle branch block & left posterior fascicular block \\
intraventricular conduction block & complete right bundle branch block \\
inferoposterolateral myocardial infarction & ST-T changes compatible with ventricular aneurysm \\
left atrial overload/enlargement & ischemic in inferolateral leads \\
digitalis-effect & anterior myocardial infarction \\
ischemic in anterolateral leads & subendocardial injury in lateral leads \\
subendocardial injury in inferolateral leads & subendocardial injury in inferior leads \\
biatrial hypertrophy & second degree AV block \\
left ventricular hypertrophy & lateral myocardial infarction \\
ischemic in anteroseptal leads & anteroseptal myocardial infarction \\
electrolytic disturbance or drug & ST-T changes \\
\hline
\rowcolor{gray!10}
\multicolumn{2}{c}{\textbf{Waveform}} \\
\hline
non-specific ST changes & low QRS voltage in chest leads \\
low QRS voltage in limb leads & Q waves present \\
low QRS voltage in left chest leads & long QTc-interval \\
long QT-interval & digitalis-effect \\
non-specific ST elevation & high QRS voltage in left ventricular \\
low QRS voltage in chest and limb leads & non-specific ST depression \\
low amplitude T-waves & ventricular premature complex \\
non-diagnostic T abnormalities & inverted T-waves \\
short PR interval & atrial premature complex \\
\hline
\rowcolor{gray!10}
\multicolumn{2}{c}{\textbf{Rhythm}} \\
\hline
sinus bradycardia & supraventricular tachycardia \\
sinus rhythm & sinus tachycardia \\
atrial bradycardia & atrial fibrillation \\
sinus arrhythmia & normal functioning artificial pacemaker \\
atrial flutter & trigeminal pattern \\
paroxysmal supraventricular tachycardia & bigeminal pattern \\
\hline
\rowcolor{gray!10}
\multicolumn{2}{c}{\textbf{Miscellaneous}} \\
\hline
burst noise & old stage of myocardial infarction\\
static noise & early stage of myocardial infarction\\
baseline drift & middle stage of myocardial infarction\\
electrodes problems & left axis deviation\\
ventricular extrasystoles & right axis deviation\\
supraventricular extrasystoles & \\
\bottomrule
\end{tabular}
\end{adjustbox}
\end{table*}

\begin{table*}[ht]
\centering
\renewcommand{\arraystretch}{1.2}
\caption{Ablation study on training stages for HeartcareGPT-3.8B on \textit{Closed-QA} tasks from Heartcare-Bench\textsuperscript{S} and Heartcare-Bench\textsuperscript{I}.}
\label{tab:stage_closed}
\begin{adjustbox}{width=0.8\textwidth,center}
\begin{tabular}{cc|cccccccccc}
\toprule
\multirow{2}{*}{\textbf{\makecell{Training\\Beat}}} & \multirow{2}{*}{\textbf{\makecell{Warming\\up}}} &  \multicolumn{4}{c}{\textbf{Heartcare-Bench\textsuperscript{S}}} & \multicolumn{4}{c}{\textbf{Heartcare-Bench\textsuperscript{I}}} & \multirow{2}{*}{\textbf{Avg.}} \\
\cmidrule(lr){3-6}
\cmidrule(lr){7-10}
& & \textbf{Diagnosis} & \textbf{Waveform} & \textbf{Rhythm} & \textbf{Misc.} & \textbf{Diagnosis} & \textbf{Waveform} & \textbf{Rhythm} & \textbf{Misc.} & \\
\midrule
- & - & 37.70 & 53.06 & 45.71 & 63.22 & 43.17 & 49.64 & 40.41 & 34.54 & 45.93 \\
- & \faCheckCircle & 45.82 & 64.37 & 57.28 & \underline{72.25} & 51.86 & 60.91 & 50.19 & 44.69 & 55.92 \\
\faCheckCircle & - & \underline{74.13} & \underline{87.10} & \underline{75.86} & 71.90 & \underline{82.03} & \underline{83.74} & \underline{71.62} & \underline{65.38} & \underline{76.47} \\
\rowcolor{lightblue}
\faCheckCircle & \faCheckCircle  & \textbf{81.95} & \textbf{95.94} & \textbf{82.79} & \textbf{79.84} & \textbf{87.85} & \textbf{92.21} & \textbf{79.25} & \textbf{67.80} & \textbf{83.33} \\
\bottomrule
\end{tabular}
\end{adjustbox}

\vspace{5mm}
\caption{Ablation study on training stages for HeartcareGPT-3.8B on \textit{Open-QA} tasks from Heartcare-Bench\textsuperscript{S} and Heartcare-Bench\textsuperscript{I}.}
\label{tab:stage_open}
\begin{adjustbox}{width=0.98\textwidth,center}
\begin{tabular}{cc|cccccccccccccccc}
\toprule
\multirow{3}{*}{\textbf{\makecell{Training\\Beat}}} & \multirow{3}{*}{\textbf{\makecell{Warming\\up}}} &
\multicolumn{8}{c}{\textbf{Heartcare-Bench\textsuperscript{S}}} & \multicolumn{8}{c}{\textbf{Heartcare-Bench\textsuperscript{I}}} \\
\cmidrule(lr){3-10}
\cmidrule(lr){11-18}
& & \multicolumn{2}{c}{\textbf{Diagnosis}} 
& \multicolumn{2}{c}{\textbf{Waveform}} 
& \multicolumn{2}{c}{\textbf{Rhythm}} 
& \multicolumn{2}{c}{\textbf{Misc.}} 
& \multicolumn{2}{c}{\textbf{Diagnosis}} 
& \multicolumn{2}{c}{\textbf{Waveform}} 
& \multicolumn{2}{c}{\textbf{Rhythm}} 
& \multicolumn{2}{c}{\textbf{Misc.}} \\
\cmidrule(lr){3-4} 
\cmidrule(lr){5-6} 
\cmidrule(lr){7-8} 
\cmidrule(lr){9-10} 
\cmidrule(lr){11-12} 
\cmidrule(lr){13-14} 
\cmidrule(lr){15-16} 
\cmidrule(lr){17-18}
& & \textbf{F1-Bio} & \textbf{Rouge-L} 
& \textbf{F1-Bio} & \textbf{Rouge-L} 
& \textbf{F1-Bio} & \textbf{Rouge-L} 
& \textbf{F1-Bio} & \textbf{Rouge-L} 
& \textbf{F1-Bio} & \textbf{Rouge-L} 
& \textbf{F1-Bio} & \textbf{Rouge-L} 
& \textbf{F1-Bio} & \textbf{Rouge-L} 
& \textbf{F1-Bio} & \textbf{Rouge-L} \\
\midrule
- & - & 50.32 & 23.15 & 57.84 & 28.97 & 49.35 & 25.42 & 60.27 & 24.32 & 53.81 & 27.19 & 55.04 & 25.96 & 50.91 & 25.30 & 56.67 & 26.18 \\
- & \faCheckCircle & 61.18 & \underline{31.84} & 61.23 & \textbf{36.95} & 66.59 & \underline{33.98} & 58.72 & \underline{30.21} & 52.84 & 28.12 & 53.47 & \underline{32.87} & 61.12 & \underline{32.45} & 55.78 & \textbf{31.55} \\
\faCheckCircle & - & \underline{66.84} & 28.21 & \underline{69.42} & 32.89 & \underline{75.23} & 33.12 & \underline{63.45} & 28.34 & \underline{60.71} & \textbf{31.59} & \underline{69.12} & 29.75 & \underline{73.67} & 31.34 & \textbf{64.27} & 28.09 \\
\rowcolor{lightblue}
\faCheckCircle & \faCheckCircle & \textbf{68.53} & \textbf{32.27} & \textbf{72.74} & \underline{35.38} & \textbf{78.63} & \textbf{36.42} & \textbf{65.84} & \textbf{30.97} & \textbf{63.17} & \underline{29.83} & \textbf{70.92} & \textbf{33.57} & \textbf{75.36} & \textbf{34.69} & \underline{61.53} & \underline{28.24}\\
\bottomrule
\end{tabular}
\end{adjustbox}

\vspace{5mm}
\caption{Ablation study on training stages for HeartcareGPT-3.8B on \textit{Comparison-QA} tasks from Heartcare-Bench\textsuperscript{C}.}
\label{tab:stage_comparsion}
\begin{adjustbox}{width=0.6\textwidth,center}
\begin{tabular}{cc|ccccccc}
\toprule
\multirow{2}{*}{\textbf{\makecell{Training\\Beat}}}
& \multirow{2}{*}{\textbf{\makecell{Warming\\up}}}
& \multicolumn{2}{c}{\textbf{S--S}}
& \multicolumn{2}{c}{\textbf{I--I}}
& \multicolumn{2}{c}{\textbf{S--I}}
& \multirow{2}{*}{\textbf{Avg.}} \\
\cmidrule(lr){3-4} \cmidrule(lr){5-6} \cmidrule(lr){7-8}
& 
& \textbf{Cons.} & \textbf{Irr.}
& \textbf{Cons.} & \textbf{Irr.}
& \textbf{Cons.} & \textbf{Irr.}
& \\
\midrule
- & - & 44.61 & 45.18 & 47.52 & 51.44 & 49.68 & 52.12 & 48.43 \\
- & \faCheckCircle & 48.55 & 50.22 & 53.18 & 55.01 & 48.87 & 51.23 & 51.18 \\
\faCheckCircle & - & \underline{63.77} & \textbf{70.44} & \underline{68.45} & \underline{69.38} & \underline{72.14} & \underline{77.29} & \underline{70.25} \\
\rowcolor{lightblue}
\faCheckCircle & \faCheckCircle & \textbf{66.40} & \underline{67.47} & \textbf{69.88} & \textbf{75.19} & \textbf{78.71} & \textbf{78.83} & \textbf{72.74}\\
\bottomrule
\end{tabular}
\end{adjustbox}

\vspace{5mm}
\caption{Ablation study on training stages for HeartcareGPT-3.8B on \textit{Report Generation} tasks from Heartcare-Bench\textsuperscript{S} and Heartcare-Bench\textsuperscript{I}.}
\label{tab:stage_report}
\begin{adjustbox}{width=0.6\textwidth,center}
\begin{tabular}{cc|cccccc}
\toprule
\multirow{2}{*}{\textbf{\makecell{Training\\Beat}}} 
& \multirow{2}{*}{\textbf{\makecell{Warming\\up}}} 
& \multicolumn{3}{c}{\textbf{Heartcare-Bench\textsuperscript{S}}} 
& \multicolumn{3}{c}{\textbf{Heartcare-Bench\textsuperscript{I}}} \\
\cmidrule(lr){3-5} \cmidrule(lr){6-8}
& & \textbf{Score\textsuperscript{GPT}} & \textbf{F1-Rad} & \textbf{Rouge-L}
& \textbf{Score\textsuperscript{GPT}} & \textbf{F1-Rad} & \textbf{Rouge-L} \\
\midrule
- & - & 55.12 & 17.11 & 26.05 & 60.32 & 19.57 & 30.27 \\
- & \faCheckCircle & \textbf{65.21} & 20.11 & 29.64 & \underline{68.02} & 21.14 & 30.89 \\
\faCheckCircle & - & \underline{64.33} & \underline{22.12} & \underline{32.20} & 66.29 & \underline{22.78} & \underline{35.43} \\
\rowcolor{lightblue}
\faCheckCircle & \faCheckCircle & 61.29 & \textbf{26.84} & \textbf{34.39} & \textbf{78.50} & \textbf{23.10} & \textbf{38.68}\\
\bottomrule
\end{tabular}
\end{adjustbox}
\end{table*}
\begin{table*}[ht]
\centering
\renewcommand{\arraystretch}{1.2}
\caption{Ablation study on multimodal integration for HeartcareGPT-3.8B on \textit{Closed-QA} tasks from proposed Heartcare-Bench\textsuperscript{S} and Heartcare-Bench\textsuperscript{I}. \textit{Seg.} = Image segmentation.}
\label{tab:multi_closed}
\begin{adjustbox}{width=0.8\textwidth,center}
\begin{tabular}{ccc|cccccccccc}
\toprule
\multirow{2}{*}{\textbf{Signal}} & \multirow{2}{*}{\textbf{Image}} & \multirow{2}{*}{\textbf{Seg.}} &
\multicolumn{4}{c}{\textbf{Heartcare-Bench\textsuperscript{S}}} &
\multicolumn{4}{c}{\textbf{Heartcare-Bench\textsuperscript{I}}} &
\multirow{2}{*}{\textbf{Avg.}} \\
\cmidrule(lr){4-7}
\cmidrule(lr){8-11}
& & & \textbf{Diagnosis} & \textbf{Waveform} & \textbf{Rhythm} & \textbf{Misc.} 
& \textbf{Diagnosis} & \textbf{Waveform} & \textbf{Rhythm} & \textbf{Misc.} 
& \\
\midrule
- & \faCheckCircle & \faCheckCircle & 74.83 & 87.12 & 76.46 & 73.42 & 80.91 & 88.05 & 69.43 & 53.36 & 75.45 \\
\faCheckCircle & - & - & 72.61 & 85.37 & 74.84 & 63.92 & 75.46 & 86.19 & 71.43 & 56.18 & 73.25 \\
\faCheckCircle & \faCheckCircle & - & \underline{79.12} & \underline{93.26} & \underline{81.64} & \underline{76.43} & \underline{84.98} & \underline{87.65} & \underline{77.43} & \underline{62.84} & \underline{80.42} \\
\rowcolor{lightblue}
\faCheckCircle & \faCheckCircle & \faCheckCircle & \textbf{81.95} & \textbf{95.94} & \textbf{82.79} & \textbf{79.84} & \textbf{87.85} & \textbf{92.21} & \textbf{79.25} & \textbf{67.80} & \textbf{83.33} \\
\bottomrule
\end{tabular}
\end{adjustbox}

\vspace{5mm}
\caption{Ablation study on multimodal integration for HeartcareGPT-3.8B on \textit{Open-QA} tasks from our proposed Heartcare-Bench\textsuperscript{S} and Heartcare-Bench\textsuperscript{I}.}
\label{tab:multi_open}
\begin{adjustbox}{width=0.98\textwidth,center}
\begin{tabular}{ccc|cccccccccccccccc}
\toprule
\multirow{3}{*}{\textbf{Signal}} & \multirow{3}{*}{\textbf{Image}} & \multirow{3}{*}{\textbf{Seg.}} &
\multicolumn{8}{c}{\textbf{Heartcare-Bench\textsuperscript{S}}} & 
\multicolumn{8}{c}{\textbf{Heartcare-Bench\textsuperscript{I}}} \\
\cmidrule(lr){4-11}
\cmidrule(lr){12-19}
& & & 
\multicolumn{2}{c}{\textbf{Diagnosis}} 
& \multicolumn{2}{c}{\textbf{Waveform}} 
& \multicolumn{2}{c}{\textbf{Rhythm}} 
& \multicolumn{2}{c}{\textbf{Misc.}} 
& \multicolumn{2}{c}{\textbf{Diagnosis}} 
& \multicolumn{2}{c}{\textbf{Waveform}} 
& \multicolumn{2}{c}{\textbf{Rhythm}} 
& \multicolumn{2}{c}{\textbf{Misc.}} \\
\cmidrule(lr){4-5} 
\cmidrule(lr){6-7} 
\cmidrule(lr){8-9} 
\cmidrule(lr){10-11} 
\cmidrule(lr){12-13} 
\cmidrule(lr){14-15} 
\cmidrule(lr){16-17} 
\cmidrule(lr){18-19}
& & & \textbf{F1-Bio} & \textbf{Rouge-L} 
& \textbf{F1-Bio} & \textbf{Rouge-L} 
& \textbf{F1-Bio} & \textbf{Rouge-L} 
& \textbf{F1-Bio} & \textbf{Rouge-L} 
& \textbf{F1-Bio} & \textbf{Rouge-L} 
& \textbf{F1-Bio} & \textbf{Rouge-L} 
& \textbf{F1-Bio} & \textbf{Rouge-L} 
& \textbf{F1-Bio} & \textbf{Rouge-L} \\
\midrule
- & \faCheckCircle & \faCheckCircle & 36.26 & 19.91 & 59.33 & 31.04 & 62.47 & 29.30 & 49.72 & 24.16 & 55.39 & 25.47 & 58.84 & 27.69 & 59.65 & 28.94 & 47.42 & 22.92 \\
\faCheckCircle & - & - & 38.74 & 21.56 & 51.62 & 26.83 & 64.29 & \underline{30.84} & 36.67 & 23.05 & 43.68 & 22.19 & 48.56 & 24.71 & 50.97 & 25.64 & 32.12 & 20.14 \\
\faCheckCircle & \faCheckCircle & - & \underline{58.18} & \underline{29.47} & \underline{70.79} & \underline{33.56} & \underline{75.91} & 29.28 & \underline{63.87} & \underline{30.29} & \underline{62.43} & \textbf{29.96} & \underline{69.71} & \underline{32.65} & \underline{73.48} & \underline{32.61} & \underline{60.28} & \textbf{31.34} \\
\rowcolor{lightblue}
\faCheckCircle & \faCheckCircle & \faCheckCircle & \textbf{68.53} & \textbf{32.27} & \textbf{72.74} & \textbf{35.38} & \textbf{78.63} & \textbf{36.42} & \textbf{65.84} & \textbf{30.97} & \textbf{63.17} & \underline{29.83} & \textbf{70.92} & \textbf{33.57} & \textbf{75.36} & \textbf{34.69} & \textbf{61.53} & \underline{28.24} \\
\bottomrule
\end{tabular}
\end{adjustbox}

\vspace{5mm}
\caption{Ablation study on multimodal integration for HeartcareGPT-3.8B on \textit{Comparison-QA} tasks from Heartcare-Bench\textsuperscript{C}.}
\label{tab:multi_comparsion}
\begin{adjustbox}{width=0.6\textwidth,center}
\begin{tabular}{ccc|ccccccc}
\toprule
\multirow{2}{*}{\textbf{Signal}} & \multirow{2}{*}{\textbf{Image}} & \multirow{2}{*}{\textbf{Seg.}}
& \multicolumn{2}{c}{\textbf{S--S}}
& \multicolumn{2}{c}{\textbf{I--I}}
& \multicolumn{2}{c}{\textbf{S--I}}
& \multirow{2}{*}{\textbf{Avg.}} \\
\cmidrule(lr){4-5} \cmidrule(lr){6-7} \cmidrule(lr){8-9}
& & & \textbf{Cons.} & \textbf{Irr.}
& \textbf{Cons.} & \textbf{Irr.}
& \textbf{Cons.} & \textbf{Irr.}
& \\
\midrule
- & \faCheckCircle & \faCheckCircle & 48.23 & 57.79 & 56.91 & 54.74 & 60.12 & 53.87 & 55.28 \\
\faCheckCircle & - & - & 51.46 & 61.32 & 53.65 & 52.83 & 64.25 & 67.01 & 58.42 \\
\faCheckCircle & \faCheckCircle & - & \underline{59.64} & \underline{63.72} & \underline{62.58} & \underline{65.84} & \underline{67.61} & \underline{71.51} & \underline{65.15} \\
\rowcolor{lightblue}
\faCheckCircle & \faCheckCircle & \faCheckCircle & \textbf{66.40} & \textbf{67.47} & \textbf{69.88} & \textbf{75.19} & \textbf{78.71} & \textbf{78.83} & \textbf{72.74}\\
\bottomrule
\end{tabular}
\end{adjustbox}

\vspace{5mm}
\caption{Ablation study on multimodal integration for HeartcareGPT-3.8B on \textit{Report Generation} tasks from Heartcare-Bench\textsuperscript{S} and Heartcare-Bench\textsuperscript{I}.}
\label{tab:multi_report}
\begin{adjustbox}{width=0.6\textwidth,center}
\begin{tabular}{ccc|cccccc}
\toprule
\multirow{2}{*}{\textbf{Signal}} & \multirow{2}{*}{\textbf{Image}} & \multirow{2}{*}{\textbf{Seg.}}
& \multicolumn{3}{c}{\textbf{Heartcare-Bench\textsuperscript{S}}} 
& \multicolumn{3}{c}{\textbf{Heartcare-Bench\textsuperscript{I}}} \\
\cmidrule(lr){4-6} \cmidrule(lr){7-9}
& & & \textbf{Score\textsuperscript{GPT}} & \textbf{F1-Rad} & \textbf{Rouge-L}
& \textbf{Score\textsuperscript{GPT}} & \textbf{F1-Rad} & \textbf{Rouge-L} \\
\midrule
- & \faCheckCircle & \faCheckCircle & 58.46 & 16.83 & 33.29 & 66.28 & 20.07 & 35.79 \\
\faCheckCircle & - & - & \underline{61.92} & 19.34 & 30.18 & 67.74 & \underline{21.99} & 32.32 \\
\faCheckCircle & \faCheckCircle & - & \textbf{64.37} & \underline{23.76} & \underline{33.84} & \underline{70.13} & 21.84 & \underline{34.49} \\
\rowcolor{lightblue}
\faCheckCircle & \faCheckCircle & \faCheckCircle & 61.29 & \textbf{26.84} & \textbf{34.39} & \textbf{78.50} & \textbf{23.10} & \textbf{38.68}\\
\bottomrule
\end{tabular}
\end{adjustbox}
\end{table*}

\definecolor{titlebar}{RGB}{200,200,200}
\definecolor{boxbg}{RGB}{255,255,255}

\begin{figure*}[ht]
\centering
\begin{adjustbox}{width=0.98\textwidth}
\begin{tcolorbox}[
enhanced,
colback=boxbg,
colframe=black,
arc=0pt,
outer arc=0pt,
boxrule=1pt,
toprule=1.5pt,
bottomrule=1pt,
leftrule=1pt,
rightrule=1pt,
titlerule=0pt,
title={\color{black}\large\textbf{Closed-QA Generation Prompt of Diagnosis}},
fonttitle=\bfseries,
attach boxed title to top left={xshift=5mm, yshift=-3mm},
boxed title style={
colback=titlebar,
colframe=titlebar,
arc=0pt,
outer arc=0pt,
boxrule=0pt,
toprule=0pt,
bottomrule=0pt,
leftrule=0pt,
rightrule=0pt,
}
]

\vspace{5mm}
\textbf{System Prompt:} 

You will be given structured ECG-related variables (age, gender, and a diagnosis-probability JSON). 
Your task is to generate exactly one JSON object that represents a single \textbf{Diagnosis Closed-QA} item. 
The output must contain the fields "type", "question", and "answer", and must follow all rules 
specified below. No additional commentary, explanations, or extra text is allowed.

\vspace{5mm}

\textbf{Instruction:}

\begin{itemize}
\item \textbf{Age:} \{Integer\}
\item \textbf{Gender:} \{Male / Female / Other\}
\item \textbf{Diagnosis Type}: \{DIAGNOSIS\_TYPE\_JSON\}\\
Each key of the diagnosis type JSON is a possible diagnosis and value is a probability (0--100). For example:
\begin{verbatim}
{
    "Inferior myocardial infarction": 95.0,
    "Atrial fibrillation": 30.0
}
\end{verbatim}
If no diagnoses are present, it is an empty JSON.

\item \textbf{Rules for Using Probabilities:}
\begin{enumerate}
\item If there is at least one diagnosis \textbf{>= 60}, choose the highest-probability diagnosis as the correct option.
\item If all diagnosis \textbf{< 60}, and the diagnosis type JSON is only composed of low-probability values or empty, then the correct answer must be one "Normal" expression.
\item A "Normal" option may be included when it is not the correct answer, and its wording can follow any standard clinical phrasing indicating normal findings (e.g., expressions such as \textit{"Normal ECG"}, "\textit{Within normal limits"}, or equivalent variants).
\item Distractors must be clinically plausible but must not overlap with any actual diagnosis in the diagnosis type JSON.
\end{enumerate}

\item \textbf{Requirements:}
\begin{enumerate}
\item The \textbf{"question"} field must be a single string and must mention both the patient’s age and gender. The phrasing should vary naturally. For example:
\begin{itemize}
\item \textit{"Based on the ECG signal of a 65-year-old male patient, ..."}
\item \textit{"For the ECG of a 70-year-old female, ..."}
\item "\textit{This ECG from a 55-year-old male patient indicates ..."}
\item \textit{"Considering the ECG tracing of a 60-year-old patient (female), ..."}
\end{itemize}
\item The question text must include ONLY four options labeled "A: ...; B: ...; C: ...; D: ..." (separated by semicolons), e.g., \textit{"Based on the ECG signal of a 65-year-old male patient, Which diagnosis is most likely? A: Atrial fibrillation; B: No abnormality detected; C: Inferior myocardial infarction; D: Complete right bundle branch block"}.
\item The \textbf{"answer"} field must be exactly the option label plus content, e.g., \textit{"A: Atrial fibrillation"}.
\item The correct answer must not consistently appear in the same option position; its placement among A--D should be randomized.
\item The output must be must be in the form of JSON:

\begin{verbatim}
{
    "type": "DiagnosisClosedQA",
    "question": ...,
    "answer": ...,
}
\end{verbatim}
\end{enumerate}
\end{itemize}

\end{tcolorbox}
\end{adjustbox}
\caption{Prompt for QA pairs generation in \textit{Closed-QA} tasks, diagnosis dimension.}
\label{fig:qa_prompt}
\end{figure*}
\definecolor{titlebar}{RGB}{200,200,200}
\definecolor{boxbg}{RGB}{255,255,255}

\begin{figure*}[ht]
\centering
\begin{adjustbox}{width=0.95\textwidth}
\begin{tcolorbox}[
enhanced,
colback=boxbg,
colframe=black,
arc=0pt,
outer arc=0pt,
boxrule=1pt,
toprule=1.5pt,
bottomrule=1pt,
leftrule=1pt,
rightrule=1pt,
titlerule=0pt,
title={\color{black}\large\textbf{Evaluation Prompt}},
fonttitle=\bfseries,
attach boxed title to top left={xshift=5mm, yshift=-3mm},
boxed title style={
colback=titlebar,
colframe=titlebar,
arc=0pt,
outer arc=0pt,
boxrule=0pt,
toprule=0pt,
bottomrule=0pt,
leftrule=0pt,
rightrule=0pt,
}
]

\vspace{5mm}
\textbf{System Prompt:} 

You are a professional cardial expert. The diagnostic accuracy of the generated report was judged according to the reference report. There are 17 evaluation indicators, and the calculation method and examples of each indicator are given below. Please compare the generated report with the reference report and score strictly according to the evaluation criteria.

\vspace{5mm}

\textbf{Instruction:}

\begin{itemize}

\item \textbf{Reference Report:} \{REFERENCE\_REPORT\}

\item \textbf{Generated Report:} \{GENERATED\_REPORT\}

\item \textbf{Evaluation Criteria:}
\begin{enumerate}
\item Completeness of abnormal features mentioned (higher=more complete): \textbf{10},
\item Completeness of key diagnoses included (higher=more complete): \textbf{10},
\item Absence of critical diagnostic errors (higher=better): \textbf{8},
\item[] $\dots$
\addtocounter{enumi}{13}
\item Whether wording is appropriate, avoiding absolute expressions: \textbf{5}
\end{enumerate}

\item \textbf{Requirements:}

\begin{enumerate}

\item Score each item in the criteria above from 0 to 100 based on comparison with the reference report.

\begin{itemize}
\item A score \textbf{from 90 to 100} indicates full compliance with the description;
\item A score \textbf{from 80 to 89} indicates substantial compliance with the description;
\item A score \textbf{from 60 to 79} indicates partial non-compliance with certain aspects;
\item A score \textbf{below 60} indicates complete non-compliance.
\end{itemize}

\item Calculate weighted dimension scores: \texttt{score\_i $\times$ weight\_i}.

\item The final total score is the sum of all weighted dimension scores:\\
\texttt{total\_score = sum(score\_i $\times$ weight\_i) / sum(weight\_i))}.

\item The output must be must be in the form of JSON:

\begin{verbatim}
{
    "item_scores": {
        "1": score_1, "2": score_2, …, "17": score_17 
    },
    "total_score": total_score
}
\end{verbatim}

\end{enumerate}

\end{itemize}

\end{tcolorbox}
\end{adjustbox}
\caption{Prompt for report evaluation.}
\label{fig:eval_prompt}
\end{figure*}

\end{document}